\documentclass[lettersize,journal]{IEEEtran}
\usepackage{amsmath,amsfonts}
\usepackage{algorithmic}
\usepackage{algorithm}
\usepackage{array}
\usepackage[caption=false,font=normalsize,labelfont=sf,textfont=sf]{subfig}
\usepackage{textcomp}
\usepackage{stfloats}
\usepackage{url}
\usepackage{verbatim}
\usepackage{graphicx}
\usepackage{cite}
\usepackage{color}
\usepackage{caption}
\usepackage{tabularx}
\usepackage{booktabs}
\usepackage[table]{xcolor}
\usepackage{multirow}
\usepackage[numbers]{natbib}
\usepackage{titlesec}

\usepackage{tabularx}
\usepackage[table]{xcolor}
\usepackage{booktabs}
\usepackage{hyperref}
\usepackage{makecell}

\usepackage{threeparttable}
\usepackage{booktabs}

\titleformat{\subsubsubsection}[runin]{\normalfont\bfseries}{\thesubsubsubsection}{1em}{}
\titlespacing{\subsubsubsection}{0pt}{\parskip}{-\parskip}
\newcommand{\subsubsubsection}{\paragraph}
\hypersetup{
  colorlinks=true,
  linkcolor=black,
  urlcolor=blue,
  citecolor=black,
}

\newcommand{\ignore}[1]{}

\newcommand{\eg}{\emph{e.g., }}

\usepackage{etoolbox}
\usepackage{flushend}
\usepackage{lettrine}

\usepackage{color,xcolor}

\begin{document}
\title{Evolutionary Reinforcement Learning: A Systematic Review and Future Directions}

\author{
    Yuanguo Lin,
    Fan Lin,
    Guorong Cai,
    Hong Chen$^{*}$,
    Lixin Zou,
    Pengcheng Wu

    \IEEEcompsocitemizethanks{
        \IEEEcompsocthanksitem Y. Lin is with the School of Computer Engineering, Jimei University, China, and also with the School of Informatics, Xiamen University, China. Email: xdlyg@jmu.edu.cn.
        \IEEEcompsocthanksitem F. Lin is with the School of Informatics, Xiamen University, China. Email: iamafan@xmu.edu.cn.
        \IEEEcompsocthanksitem G. Cai is with the School of Computer Engineering, Jimei University, China. Email: guorongcai.jmu@gmail.com.
        \IEEEcompsocthanksitem H. Chen is with the Institute of Systems Science, National University of Singapore, Singapore. Email: chenhong@u.nus.edu.
        \IEEEcompsocthanksitem L. Zou is with the school of Cyber Science and Engineering, Wuhan University, Wuhan, China. Email: zoulixin@whu.edu.cn.
        \IEEEcompsocthanksitem P. Wu is with the Joint NTU-UBC Research Centre of Excellence in Active Living for the Elderly (LILY), Nanyang Technological University, Singapore. Email: pengchengwu@ntu.edu.sg.
        
        \IEEEcompsocthanksitem $^{*}$ Corresponding author
    }
}

\markboth{Journal of \LaTeX\ Class Files,~Vol.~14, No.~8, August~2021}%
{Shell \MakeLowercase{\textit{et al.}}: A Sample Article Using IEEEtran.cls for IEEE Journals}


\maketitle

\begin{abstract}
In response to the limitations of reinforcement learning and evolutionary algorithms (EAs) in complex problem-solving, Evolutionary Reinforcement Learning (EvoRL) has emerged as a synergistic solution. EvoRL integrates EAs and reinforcement learning, presenting a promising avenue for training intelligent agents. This systematic review firstly navigates through the technological background of EvoRL, examining the symbiotic relationship between EAs and reinforcement learning algorithms. We then delve into the challenges faced by both EAs and reinforcement learning, exploring their interplay and impact on the efficacy of EvoRL. Furthermore, the review underscores the need for addressing open issues related to scalability, adaptability, sample efficiency, adversarial robustness, ethic and fairness within the current landscape of EvoRL. Finally, we propose future directions for EvoRL, emphasizing research avenues that strive to enhance self-adaptation and self-improvement, generalization, interpretability, explainability, and so on. Serving as a comprehensive resource for researchers and practitioners, this systematic review provides insights into the current state of EvoRL and offers a guide for advancing its capabilities in the ever-evolving landscape of artificial intelligence.
\end{abstract}


\section{Introduction}
\lettrine[lines=3, findent=3pt, nindent=0pt]{\textbf{R}}{einforcement} learning utilizes agents for autonomous decision-making, focusing on long-term action strategies, particularly effective in tasks like industrial automation and personalized recommendation systems \cite{kaelbling1996reinforcement, morimoto2005robust, zhao2018deep}. However, reinforcement learning faces challenges such as parameter sensitivity and sparse rewards, leading to issues in learning efficiency and adaptability \cite{de2023out, liu2021pns}. On the other hand, evolutionary algorithms (EAs), inspired by Darwin's natural selection, excel in solving complex, multi-objective problems in large solution spaces \cite{goldberg1989cenetic, coello2007evolutionary}. Despite their robust search capabilities, EAs are limited by hyper-parameter sensitivity and struggle in high-dimensional environments, affecting their optimization efficacy \cite{suri2020maximum, zhang2023reinforcement}.

Hence, Evolutionary Reinforcement Learning (EvoRL) which integrates EAs with reinforcement learning, to address the limitations of each method \cite{nilsson2021policy}. EvoRL maintains multiple policies within a population and utilizes evolutionary operations like crossover and mutation to refine these policies, enhancing the policy-making process inherent in reinforcement learning. Simultaneously, EvoRL leverages the global search capabilities of EAs for exploring the policy space and optimizing various components like agents and actions. EvoRL's core mechanism, combining the precision of policy gradients with EAs' global search, enables effective solutions in complex, high-dimensional environments \cite{khadka2018evolution, shi2020efficient}. Additionally, approaches like EGPG and CERL within EvoRL focus on collaborative efforts of multiple agents, boosting performance in intricate tasks \cite{khadka2019collaborative}.

EvoRL has been applied in various domains, demonstrating its versatility and effectiveness. For instance, EvoRL enhances sample efficiency in reinforcement learning, a crucial aspect for practical applications \cite{lu2021recruitment, franke2020sample}. In embodied intelligence, EvoRL fosters complex behavior through the integration of learning and evolution, offering new perspectives in this field \cite{gupta2021embodied}. Another significant application of EvoRL lies in quality diversity for neural control, contributing to the advancement of neural network-based control systems \cite{pierrot2020sample}. EvoRL's integration with Deep Reinforcement Learning (DRL) has been instrumental in promoting novelty search, expanding the boundaries of exploration in reinforcement learning \cite{shi2020efficient}. Furthermore, the SUPER-RL approach within EvoRL, which applies genetic soft updates to the actor-critic framework, has been shown to improve the searching efficiency of DRL \cite{marchesini2020genetic}. Early research highlighted the importance of meta-parameter selection in reinforcement learning, a concept that remains relevant in current EvoRL applications \cite{eriksson2003evolution}. Lastly, the combination of population diversity from EAs and policy gradient methods from reinforcement learning has led to novel approaches like differentiable quality diversity, further enhancing the gradient approximation capabilities in EvoRL \cite{tjanaka2022approximating}.

Recent advances in EvoRL \cite{majid2023deep, wang2021evolutionary} have demonstrated significant improvements in both reinforcement learning and EAs. From the perspective of reinforcement learning, EvoRL has been shown to notably enhance sample efficiency and expand exploratory capabilities, which are essential in addressing reinforcement learning's limitations in complex and high-dimensional problem spaces \cite{shi2020efficient, franke2020sample}. In terms of EAs, the integration with reinforcement learning techniques has resulted in more adaptive and precise evolutionary strategies \cite{eriksson2003evolution, marchesini2020genetic}. This review aims to underscore the importance of EvoRL in overcoming the inherent challenges of using reinforcement learning and EA independently, highlighting its integrated approach in complex problem-solving scenarios.

In the field of EvoRL, the surveys by \cite{sigaud2023combining,bai2023evolutionary} provide related insights. \cite{sigaud2023combining} categorizes over 45 EvoRL algorithms, primarily developed after 2017, focusing on the integration of EAs with reinforcement learning techniques, which emphasizes the mechanisms of combination rather than the experimental results. This classification provides a structured approach to understanding the field's current methodologies. \cite{bai2023evolutionary} further explores this domain, examining the intersection of EAs and reinforcement learning, and highlights its applications in complex environments. Although these surveys offer valuable insights for EvoRLs, they might benefit from incorporating a broader range of perspectives and analytical approaches, ensuring a more diverse understanding of EvoRL's potential and challenges in future explorations.

\textbf{Our contribution} In this review, the contribution made by us can be demonstrated by following four points:
\begin{itemize}
    \item \textbf{Multidimensional classification} We have taken a comprehensive look at the EvoRL field, categorizing different approaches and strategies in detail. This detailed classification covers not only EvoRL's multiple techniques and strategies, such as genetic algorithms and policy gradients, but also their application to complex problem-solving.
    \item  \textbf{Multifaceted issue comments} We provide an in-depth analysis of the challenges and limitations of current research, contains issues encountered by reinforcement learning and EAs and their corresponding EvoRL outcome methods, and provide critical insights into existing research methods.
    \item \textbf{Constructive Open Issues} This review presents a summary of emerging topics, including scalability to high-dimensional spaces, adaptability to dynamic environments, sample efficiency and data efficiency, adversarial robustness, ethic and fairness in EvoRL.
    \item \textbf{Promising Future Directions} We identify key research gaps and propose future directions for EvoRL,  highlighting the need of meta-evolutionary strategies, self-adaptation and self-improvement mechanisms, transfer learning and generalization, interpretability and explainability, as well as incorporating EvoRL within large language models.
\end{itemize}

\textbf{The remainder of this review} It begins with an overview of the technological background used in EvoRL, followed by an investigation of EvoRL's applications and algorithms based on existing literature. The review then analyses the challenges faced by both EAs and reinforcement learning, along with the solutions provided by EvoRL. Finally, it concludes with the discussion on the open issues and future directions in the field of EvoRL.
\section{Background}
\label{Background}
In the literal sense, EvoRL combines EAs with reinforcement learning. To offer a comprehensive description of EvoRL, this section will provide brief introductions to both EAs and reinforcement learning.

\subsection{Evolutionary Algorithms}
\label{EA}
Evolutionary algorithms do not refer to one specific algorithm, but the generic term of a series of sub-algorithms that are inspired by natural selection and the principle of genetics. Those sub-algorithms solve complex problems by imitating the process of biological evolution. Therefore, EAs can be considered as a general concept, that consists of multiple optimization algorithms which utilize the mechanism of biological evolution \cite{vikhar2016evolutionary}.

In this paper, five areas of EAs will be introduced, which are evolutionary strategy, genetic algorithm, cross-entropy method (CEM), population-based training (PBT) and other EAs.

\textbf{Evolutionary strategy} was first proposed by Rechenberg \cite{hoffmeister1990genetic,vent1975rechenberg}. Evolutionary strategy optimize the solution through population generation and mutation based on principles of biological evolution. There is an adaptive adjustment of mutation step size in this algorithm, which enables the method to search effectively in the solution space. Following is core formula of evolutionary strategy, which is applied on the adjustment of mutation step size:
\begin{equation}
    \begin{split}
    &\sigma^{\prime}_j = \sigma_j \cdot e^{(\mathcal{N}(0,1)-\mathcal{N}_j(0,1))},\\
    \end{split}
\end{equation}
where $\sigma^{\prime}_j, \sigma_j$ denote the mutation step size after and before adjustment respectively, $e^{(\mathcal{N}(0,1)-\mathcal{N}_j(0,1))}$ represents the adjust factor of normal distribution random number, which used for increasing or decreasing step size. Evolutionary strategy is especially suitable for solving issues of sparse reward and policy search in reinforcement learning due to its strong capability of searching on large scale\cite{bai2023evolutionary}.

\textbf{Genetic algorithm} is one of the most famous EAs. There are three essential operators which are selection, mutation, and crossover \cite{zhu2023survey}. Population is also a core concept in genetic algorithm, each individual in it represents a potential solution, which will be evaluated through fitness function\cite{slowik2020evolutionary}. Different from evolutionary strategy, genetic algorithm focuses on the crossover as the main mechanism of exploration, while evolutionary strategy tends to rely on mutation\cite{hoffmeister1990genetic}. Genetic algorithm was usually applied to deal with the problem of hyper-parameter tuning in reinforcement learning.

\textbf{Cross-entropy method} defines an accurate mathematical framework, that explores optimal solutions based on advanced simulation theory. It starts by randomizing the issue using a series of probability density functions and solves the associated stochastic problem through adaptive updating \cite{ho2010cross}. The key concept of CEM contains elite solution-based probability density function's parameters updating and global searching ability enhancement by applying mutation. Following is the core formula of CEM \cite{botev2013cross}:
\begin{equation}
    \mathbf{v^*}=\underset{\mathbf{v}}{\arg\max}\mathbb{E}_\mathbf{W}\mathbf{I}_{{S(\mathbf{X})\geq\gamma}}\log f(\mathbf{X};\mathbf{v})\frac{f(\mathbf{X};\mathbf{u})}{f(\mathbf{X};\mathbf{w})},
\end{equation}
where $S(\mathbf{X})$ is a performance function used to evaluate random variable $\mathbf{X}$, $f(\mathbf{X};\mathbf{v})$ denotes a parameterized probability density function, $\mathbf{v}$ is a parameter. This equation aims to find a $\mathbf{v^*}$ to maximize the logarithmic likelihood expectation when the event $S(\mathbf{X})\geq\gamma$ happens, $\gamma$ is a threshold. That is, seek a $\mathbf{v^*}$ to make a high-performance event easier to happen. This is the core idea of CEM, which is by minimizing cross-entropy, the sampling distribution is gradually close to the high-performance region so that the high performance is more easily sampled \cite{botev2013cross}. CEM offers a robust solution to the challenges of local optima and computational efficiency in reinforcement learning \cite{huang2021cem}.

\textbf{Population-based training} can be viewed as a type of parallel EAs, it optimizes weights and hyper-parameters of a neural network's population simultaneously \cite{jaderberg2017population}. The algorithm starts with randomly initializing a population of models, each model individual will optimize their weights independently (similar to mutation) in training iteration. The prepared individual needs to get through two stages which are "exploit" and "explore" at the population level. The former is similar to "selection", that is, replacing original model weights and hyper-parameters with a better performance model. Actually, it is a parameter transfer inside the population. The latter one is similar to "mutation", realizing the exploration of hyper-parameters space by adding noise, which provides diversity for following training \cite{jaderberg2017population}. Due to its characteristics, PBT shows an excellent ability to solve issues of hyper-parametric sensitivity and sparse rewards in reinforcement learning \cite{bai2023evolutionary,zhu2023survey}.

Apart from four algorithms introduced previously, there are several other EAs, \eg genetic programming, evolutionary computing, and random search. Although these algorithms have their own characteristics, they basically share some common principles like population-based search, iterative optimization, simulation of natural selection, and genetic mechanisms. Genetic programming is mainly applied to optimizing policy search \cite{hein2018interpretable}, evolutionary computing focuses on fitness function problem \cite{sachdeva2021maedys} while random search could be used for improving calculation efficiency of reinforcement learning \cite{shi2019fidi}.

\subsection{Reinforcement Learning}
As one of the most popular machine learning methods, reinforcement learning has attracted a lot of attention in recent years. It has shown great potential in everything from gaming to autonomous driving. Given that reinforcement learning has been extensively discussed and studied, we will not repeat its basic concepts here, but directly introduce its three main methods.

\textbf{Value function} approach evaluates each state $s$ or state-action pair $(s,a)$ by learning a value function $v_\pi(s,a)$, and guides decision making according to this \cite{lin2023comprehensive}, which can also be demonstrated as:
\begin{equation}
    \pi(s)=\underset{a}{\arg\max} v_\pi(s,a),
\end{equation}
where $\pi(s)$ denotes the policy under state $s$. It can be also called as greedy policy \cite{chen2021survey}. The core concept of this approach is to optimize the value function to predict future rewards.

\textbf{Policy gradient method} directly optimizes the policy $\pi(a|s)$ itself, aiming to maximize the expected return \cite{lin2023survey}. This method optimizes the policy by adjusting policy parameters but not the estimated value function. The policy gradient method provides a formula related to updating the policy parameters $\theta$ by gradient ascent on the expected return \cite{chen2021survey}:
\begin{equation}
    J(\pi_\theta) =\mathbb{E}_{\tau \sim \pi_\theta}\lbrack r(\tau)\rbrack= \int\pi_\theta(\tau) r(\tau) \, d\tau,
\end{equation}
where $J(\pi_\theta)$ denotes the expected return of the policy, $\tau$ is a trajectory, $r(\tau)$ represents cumulative reward after operating trajectory $\tau$, while $\mathbb{E}_{\tau \sim \pi_\theta}$ demonstrates the expectation over trajectories sampled from the policy $\pi_\theta$ \cite{chen2021survey}.

\textbf{Actor-critic algorithm} is a reinforcement learning technique used for learning optimal policies in unknown environments. Its core concept revolves around guiding the policy improvement of the actor based on the value function estimations provided by the critic \cite{lin2023comprehensive,lin2023survey}. In this algorithm, the updating rule of policy parameter $\theta$ is defined as:
\begin{equation}
    \Delta\theta=\alpha\nabla\log \pi_\theta(s,a)(r+\gamma V(s^\prime)-V(s)),
\end{equation}
where $\alpha$ represents the learning rate, $\pi_\theta(s,a)$ denotes the policy, $V(s)$ and $V(s^\prime)$ are the estimation of value function under current and next state, respectively. $r$ is the reward, while $\gamma$ represents the discount factor towards future rewards \cite{chen2021survey}.
\section{Methods of Evolutionary Reinforcement Learning}
\label{Methodologies}
This section aims to provide an in-depth discussion of the core mechanism in EvoRL algorithms, focusing on the EAs they incorporate. In this section, the components of reinforcement learning and EAs will be described in detail, to analyze their effects and advancements in the decision-making process. Besides, we will also discuss the evaluation metrics used for assessing EvoRL's performance and the corresponding benchmarks. An overview of literature is shown as Table~\ref{tab:EvoRLalgo}.

\begin{table*}[htbp]
  \centering
  \caption{Overview of Evolutionary Reinforcement Learning Algorithms with Evaluation Metrics and Benchmarks}
  \footnotesize
   \resizebox{\linewidth}{!}{
   \begin{tabular}{lllll}
    \toprule
    \textbf{Evolutionary Algorithm} & \textbf{RL Algorithm} & \textbf{Method} & \textbf{Evaluation Metric} & \textbf{Compared Algorithm} \\
    \midrule
    \multirow{8}[0]{*}{Evolutionary Strategy} & \multirow{2}[0]{*}{Value Function}  & AEES \cite{ajani2022adaptive} & Convergence, Cumulative return & A2C,SAC,DDPG,TRPO \\
          &  &   OHT-ES \cite{tang2020online}    & Asymptotic performance,learning speed & TD3 \\
\cmidrule{2-5}
          & \multirow{2}[0]{*}{Policy Gradient}  &R-R1-ES \cite{chen2019restart} & Reward & OpenAI-ES,NS-ES \\
          &  &   CCNCS \cite{yang2022evolutionary}    & Time budget & PPO,A3C \\
\cmidrule{2-5}          
          & \multirow{4}[0]{*}{Actor-Critic}  & GRL \cite{su2023evolution} & Makespan,Computation complexity & PPO,L2D \\
          &  &   ZOSPI \cite{sun2020zeroth}    & Reward & SAC,TD3,OAC,OURS \\
          &   &  EPG \cite{houthooft2018evolved}     & Return, KL & PPO \\
          &  &   ES-TD3 \cite{callaghan2023evolutionary}    & Mean,std,Median & TD3,CEM-RL \\
    \midrule
    \multirow{13}[0]{*}{Genetic Algorithm} & \multirow{4}[0]{*}{Value Function} & A-MFEA-RL \cite{martinez2021adaptive} & Time budget & SAC,PPO \\
          &  &   MAERL \cite{li2023multi}   & MAE,RMSE,MME & TD3 \\
          &  &   ERLGA \cite{zheng2023rethinking}    & Return & NA \\
          &  &   COE-RL \cite{elfwing2008co}    & Time budget,episode & NA \\
\cmidrule{2-5}          & \multirow{3}[0]{*}{Policy Gradient} & AMODE-DRL \cite{li2023scheduling} & Time budget & DDPG,DDQN \\
          &   &  MERL \cite{yang2023reducing}     & Convergence & PPO,IPG \\
          &  &   MetaPG \cite{garau2022multi}    & Entropy,Return & SAC \\
\cmidrule{2-5}          
          & \multirow{6}[0]{*}{Actor-Critic} & SI-LFC \cite{li2023evolutionary} & Frequency deviation, Generation cost & MADDPG,MATD3 \\
          &   &  ECRL \cite{hu2023evolving}     & Constraint, Reward & IPO \\
          &  &   PDERL \cite{bodnar2020proximal}    & Reward  & TD3,PPO \\
          &   &  SERL \cite{wang2022surrogate}     & Step  & DDPG \\
          &  &   QD-PG-PF \cite{pierrot2022diversity}    & Training curve & DPG \\
          &  &   ERL-Re \cite{hao2022erl}    & Return & TD3 \\
    \midrule
    \multirow{7}[0]{*}{Cross-Entropy Method} & Value Function  & QT-OPT \cite{kalashnikov2018qt}& Success rate & NA \\
\cmidrule{2-5}          
          & Policy Gradient & GRAC \cite{shao2022grac} & Return & TD3,SAC,DDPG,TRPO,CEM \\
\cmidrule{2-5}          
          & \multirow{5}[0]{*}{Actor-Critic} &  CEM-RL \cite{pourchot2018cem} & Training curve & TD3 \\
          &  & PGPS \cite{kim2020pgps}      & Return & CEM,PPO,DDPG,CERL,SAC \\
          &  & CSPC \cite{zheng2020cooperative}      & Return & SAC,PPO,CEM \\
          &  & SAC-CEPO \cite{shi2021soft}     & Return & SAC \\
          &  &  DEPRL \cite{liu2021diversity}     & Return & CEM,TD3 \\
    \midrule
    \multirow{9}[0]{*}{Population-Based Training} & \multirow{2}[0]{*}{Policy Gradient} & SBARL \cite{de2023out} & NA    & TD3 \\
          &   &  EARL \cite{wang2023evolutionary}     & NPV   & SAC \\
\cmidrule{2-5}          
          & \multirow{7}[0]{*}{Actor-Critic} & EAS-RL \cite{ma2022evolutionary} & Return & TD3 \\
          &  &  PS3-TD3 \cite{jung2020population}     & Reward & TD3 \\
          &   &  ARAC \cite{doan2019attraction}     & Return & CERL,TD3 \\
          &  &    SOS-PPO \cite{marchesini2022exploring}   & Reward & PPO \\
          & &    MERL \cite{majumdar2020evolutionary}    & Success rate & MADDPG,MATD3 \\
          &  &   EPC-MARL \cite{long2020evolutionary}    & Reward & MADDPG \\
          &  &   EMOGI \cite{shen2020generating}    & Win rate,Duration,Distance & A3C \\
    \midrule
    \multirow{8}[0]{*}{Others} & \multirow{4}[0]{*}{Value Function}  & ECRL \cite{fernandez2018parameters} & Reward & NA \\
          &  & AGPRL \cite{kamio2005adaptation}      & Q-value & NA \\
          &  & TDDQN \cite{co2021evolving}      & Q-value & DQN \\
          &  & DDQN-RS \cite{abuzekry2019comparative}      & MLHP  & DDQN \\
          \cmidrule{2-5}
          & \multirow{3}[0]{*}{Policy Gradient} & EGPRL \cite{kelly2021evolving} & Reward & NA \\
          &   &   GPRL \cite{hein2018interpretable}    & Error & NA \\
          &  &    GPFDM \cite{girgin2008feature}   & Reward & NA \\
          \cmidrule{2-5}
          & Actor-Critic & FiDi-RL \cite{shi2019fidi} & NA    & NA \\
    \bottomrule
    \end{tabular}%
    }
  \label{tab:EvoRLalgo}%
\end{table*}%

\subsection{Evolutionary Strategy}
In the expansive field of reinforcement learning, the value function method has always been one of the core research directions, mainly focusing on how to effectively estimate and optimize the expected return under a given policy. Against this background, the natural parameter search mechanism of evolutionary strategy provides a unique approach by simulating the processes of natural selection and heritable variation.

Mutation is one of the most essential operations in the evolutionary strategy, it provides the algorithm with new solution space. The operation empowers the corresponding EvoRL to effectively adapt complex learning environments. \cite{ajani2022adaptive} employs a simple but effective EvoRL algorithm called AEES, which contains two distinct, coexisting mutation strategies. Each two strategies is connected with their population subsets. That is, each subset mutates in accordance with one related mutation strategy. AEES applies cumulative return and convergence rate as evaluation metrics, and shows a better performance of the proposed model compared to A2C, SAC, and other deep reinforcement learning (DRL) methods. Compared to \cite{ajani2022adaptive}, the OHT-ES algorithm in \cite{tang2020online} more focuses on adjusting key parameters of the reinforcement learning method by evolutionary strategy, hence improving the adaptability and efficiency of the algorithm. \cite{tang2020online} proves that OHT-ES performs better than traditional DRL (\eg TD3) in learning speed.

Different from value function, the introduction of evolutionary strategy in policy gradient method provides a brand new perspective. The realization idea of R-R1-ES \cite{chen2019restart} is different from \cite{ajani2022adaptive}, R-R1-ES put special emphasis on the direct optimization of the policy itself, which applies the Gaussian distribution model $N(\theta_t,\sigma^2C_t)$ and restart mechanism to update the searching direction, where $\theta_t \in \mathbb{R}^n$ represents distribution mean, $\sigma_t>0$ denotes mutation strength, $C_t$ is a n-dimensional covariance matrix at $t$ iteration. The update rule of $C_t$ is given as:
\begin{equation}
    C_t = (1-c_{cov})I+c_{cov}p_tp_t^T,
\end{equation}
where $c_{cov} \in (0,1)$ is the changing rate of covariance matrix, $I$ denotes unit matrix, $p_t$ is a vector which represents primary search direction. The model performs better than NS-ES (Novelty Search-Evolutionary Strategy) according to reward evaluation.

Besides, the ZOSPI model \cite{sun2020zeroth} reveals the potential of combining evolutionary strategy and actor-critic algorithm. Compared to R-R1-ES \cite{chen2019restart}, ZOSPI chooses to optimize policy from global and local aspects, which both exploit the advantages of the global value function and the accuracy of policy gradient to the full. 
\begin{equation}
    \nabla_\phi J=E\lbrack\nabla_a Q_w(s_t,a)|a=\pi_{\theta_t} (s_t)+\pi_{\phi_t} (s_t)\nabla_\phi \pi_{\phi_t}(s_t)\rbrack,
\end{equation}
where $\nabla_\phi J$ is the gradient of object function $J$ to perturbation network's parameter $\phi$, $\nabla_a Q_w(s_t,a)$ represents the gradient of value function to action $a$, $\pi_{\theta_t} (s_t)$ denotes the action selected by current policy, $\pi_{\phi_t} (s_t)$ is the output of perturbation network under time step $t$ and status $s_t$, $\nabla_\phi \pi_{\phi_t}(s_t)$ denotes the gradient of perturbation network to its parameter $\phi$. The approach not only improves sample efficiency but expands the possibility of multi-modal policy learning, paving a novel trajectory for actor-critic algorithm.
\subsection{Genetic Algorithm}
Genetic algorithm-based EvoRLs are different from traditional methods, they focus on applying genetic diversity to the policy search process, which makes the algorithm find an effective and stable policy in a highly complex environment.

One of the valuable contributions of genetic algorithms is the variety of parameters and strategies provided by crossover operation. \cite{li2023multi} introduces a MAERL method that mainly focuses on parameter optimization in the processing industry. The method consists of multi-agent reinforcement learning, Graph Neural Networks (GNNs), and genetic algorithms. The ERLGA \cite{zheng2023rethinking} method discusses the combination of genetic algorithm and off-policy reinforcement learning, finally reaching a better performance on return than existing methods. Moreover, \cite{garau2022multi} proposes a MetaPG algorithm to optimize different reinforcement learning targets by multi-objective searching standards, and consider individual reinforcement learning targets via Non-dominated Sorting Genetic Algorithm II (NSGA-II).
\begin{equation}
    \begin{split}
        L^{perf}_\pi = \mathbb{E}_{(s_t,a_t,s_{t+1})~D}\lbrack \log&(\min(\pi(\Tilde{a}_{t+1}|s_{t+1})\gamma))\\
        &-\min Q_i(s_t,\Tilde{a}_t)\rbrack,    
    \end{split}
\end{equation}
where $\pi(\Tilde{a}_{t+1}|s_{t+1})$ denotes the policy probability of taking action $\Tilde{a}_{t+1}$ given a next state $s_{t+1}$, $\gamma$ represents discount factor, $Q_i(s_t,\Tilde{a}_t)$ is action value function. The formula in \cite{garau2022multi} denotes the policy loss function. MetaPG is able to improve about $3\%$ and $7\%$ in performance and generalization compared to Soft Actor-Critic (SAC) by adjusting the loss function. Simultaneously, genetic algorithms play a crucial role in fine-tuning the intricacies of policy evolution, \cite{bodnar2020proximal} develops a PDERL method to solve the scalability issues caused by simple genetic encoding in traditional algorithms. The PDERL applies the following formula to define the proximal mutation operator:
\begin{equation}
    s = \frac{1}{|A|N_M}\sum_i \| \nabla_\theta \mu_\theta(s_i)\| ^2,
\end{equation}
where $s$ represents the sensitivity of action to weight perturbation, $|A|$ is the size of action space, $N_M$ denotes the sample size used for calculating sensitivity, $\nabla_\theta \mu_\theta(s_i)$ is the gradient of policy network to its parameter $\theta$, to evaluate the sensitivity of policy changes to parameter under state $s_i$.

In addition, genetic algorithms can be applied to tackle complex reinforcement learning problems that demand extensive interaction with the environment, \cite{wang2022surrogate} proposes a SERL which contains a Surrogate-assisted controller module. The module combines genetic algorithm and actor-critic algorithm, where the genetic algorithm here is mainly used for evaluating the fitness of the genetic population. The method applies surrogate models to predict the environmental performance of individuals, which decreases the requirements of direct interaction with the environment and leads to a lower computational cost.

\subsection{Cross-Entropy Method}
In the field of multiple EvoRLs research, CEM as a core technique, mainly focuses on selecting elites to update policy distribution, so that policy evolves in a better direction. The key concept of CEM-based EvoRL is it does not rely on complex gradient calculation, but processes iterative optimization of policies by statistical methods.

The CEM-RL \cite{pourchot2018cem} is a typical example, it combines CEM with policy gradient, to balance exploration and exploitation \cite{pourchot2018cem}. \cite{shi2021soft} proposes a SAC-CEPO method combining CEM and SAC. More specifically, SAC-CEPO samples optimal policy distribution iteratively and applies it as the target of policy network update. The key formula of SAC-CEPO is:
\begin{equation}
    J(\pi) = E_{\pi} \left[ \sum_t \gamma^t r(s_t, a_t) + \alpha H(\pi(\cdot|s_t)) \right],
\end{equation}
where $J(\pi)$ represents the performance of policy $\pi$, $E_\pi$ is the expectation under the policy $\pi$, $\gamma$ denotes discount factor, $r(s_t,a_t)$ indicates the reward of action $a_t$ under state $s_t$, $\alpha$ stands for the model parameter.

Not only can CEM optimize the policy network effectively, but it can also enhance the overall decision quality and algorithmic efficiency through the statistical evolution of the value function. \cite{kalashnikov2018qt} introduces the QT-OPT method, which applies CEM to optimize the value function of reinforcement learning. QT-OPT shows a good ability on success rate evaluation compared to existing algorithms. \cite{shao2022grac} develops an algorithm that exploits CEM to seek the optimal action with maximal Q-value, called GRAC. The combination of EAs and value function makes GRAC exceed TD3, SAC, and other popular DRLs in OpenAI gym's six continuous control missions. 

CEM also demonstrates a robust capability in guiding population evolution and iteratively optimizing the entire policy space, thereby expanding its application in the field of reinforcement learning. \cite{kim2020pgps} proposes a PGPS that considers CEM as its core component. In PGPS, CEM is applied to generate the next population based on current evaluation results, to create a higher return (defined as the sum of instant rewards within a certain number of steps). PGPS performs better than multiple DRLs such as DDPG, PPO, SAC, etc. in several MuJoCo environments.
\subsection{Population-Based Training}
PBT-based EvoRL shows the potential in multiple research fields. The core concept of it is to adjust parameters and hyper-parameters of the algorithm dynamically in the training process, so that realizes a more effective and flexible learning in complicated environments.

Therefore, \cite{de2023out} introduces a parameter control strategy training method for EA and swarm intelligence (SI) algorithms called SBARL. PBT in SBARL is applied to evolve parameters and hyper-parameters in reinforcement learning, the results of the experiment demonstrate that SBARL performs better than traditional DRL TD3. EARL \cite{wang2023evolutionary} is a framework that combines EA (more specifically, PBT) and reinforcement learning. The key concept of EARL is the collaborative work of EA and reinforcement learning could facilitate the learning and evolving process. \cite{majumdar2020evolutionary} proposes a MERL algorithm, that realizes individual and team objectives by combining gradient-less and gradient-based optimizers. Among them, The policy of gradient-based will be added to the population evolving regularly, which enables EAs to leverage skills learned through training on individual-specific rewards to optimize team goals without relying on reward shaping. Besides the policy space, PBT can influence the action space in reinforcement learning, emphasizing the importance of optimizing action decisions. \cite{ma2022evolutionary} proposes an EAS-RL method, it exploits actions selected by reinforcement learning's policy to generate a population, and processes particle swarm optimization to evolve the population iteratively. The key concept of EAS-RL is choosing to optimize action space but not policy space, the definition of loss function in action space is:
\begin{equation}
    L_{evo}(\theta,\mathcal{A})=\mathbb{E}_{(s_i,a_i^e)\sim \mathcal{A}} \lbrack \| \mu_\theta(s_i)-a^e_i \| ^2\rbrack,
\end{equation}
where the state $s_i$ and the evolutionary action $a_i$ in this formula are sampled respectively from the archive $\mathcal{A}$, while $\theta$ denotes the learning parameters in the reinforcement learning policy $\mu_\theta$. The proposed model behaves better than TD3 in MuJoCo environments.

Beyond optimizing parameters and actions, PBT places specific emphasis on automatically adjusting reward functions, \eg \cite{shen2020generating} focuses on game-AI generation. Therefore, they propose a PBT-based EvoRL, EMOGI framework. EMOGI considers the reward function as a part of candidate objects, realizing the auto-adjustment of parameters by EAs. EMOGI applies multi-objective optimization to select policies with distinct behaviors, to ensure population diversity. The key process of initialization of EMOGI, that is, randomly initialize a candidate population consisting of policy parameters and reward weights:
\begin{equation}
    P=\{\pi_{\theta_1}R_{w_1},...,\pi_{\theta_n}R_{w_n}\},
\end{equation}
where $P$ denotes the population of candidates, each candidate consists of two parts, policy parameter $\pi_\theta$, and weight of the reward $R_w$. The size of the population is decided by $n$, representing the amount of candidates in the population.

\subsection{Other EAs}
The preceding chapters covered classical algorithms based on evolutionary strategies, genetic algorithms, CEM, and PBT, each demonstrating notable results in their respective application scenarios. Additionally, there are other promising EvoRL methods that, while less commonly employed, deserve attention. These include random search-based approaches designed to enhance efficiency by streamlining the search process, genetic programming methods that optimize strategies through simulation of biological genetic processes, and evolutionary computing algorithms that emphasizes the use of principles of evolutionary theory to improve the learning process.

Random search simplifies the parameter optimization process while maintaining a certain level of exploration ability. This approach can efficiently find solutions in complex tasks. DDQN-RS \cite{abuzekry2019comparative} applies random search to randomly sample individuals from the population by Gaussian distribution. Evaluate their fitness according to the reward got from one round of running in the environment. Compared to Double Deep Q-Network (DDQN), the proposed model performs better than it does in the mission of keeping the vehicle close to the center of the lane for the longest distance. 

Compared to the efficient parameter optimization discussed in \cite{abuzekry2019comparative}, genetic programming in \cite{co2021evolving} demonstrates a capability for in-depth optimization of computation graphs. In this framework, genetic programming is applied to search in computation graph space, and these graphs compute the minimal objective function required by the agent. Genetic programming improves these computation graphs by simulating the evolving process of creatures. The proposed model applies the loss function from Deep Q-Network (DQN) as a key component:
\begin{equation}
    L_{DQN}=(Q(s_t,a_t)-(r_t+\gamma\cdot\max_aQ_{targ}(s_{t+1},a)))^2,
\end{equation}
where $Q(s_t,a_t)$ indicates the Q-value under current state $s_t$ and action $a_t$, $r_t$ represents instant reward, $\gamma$ denotes discount factor, $\max_aQ_{targ}(s_{t+1},a)$ stands for the maximal expectation of Q-value under next state $s_{t+1}$. By applying this loss function, the proposed method more focuses on a more accurate estimation of the Q-value, to solve the overestimate issue. The experiment result shows that the DQN modified by genetic programming behaves better than the original DQN in Q-value estimation \cite{co2021evolving}.

Similarly, \cite{kelly2021evolving} proposes an EGPRL that applies genetic programming to search on computation graph space, finally minimizing the objective function. EGPRL allows agents could operate under multiple environments including OpenAI Gym’s Classic Control Suite. The experiment result shows that the proposed model owns a competitive generalization ability and efficiency. Different from \cite{co2021evolving} which considers accurate estimation of Q-value as its core concept, EGPRL more focuses on the hierarchical structure of memory coding and multitasking.

Besides, ECRL \cite{fernandez2018parameters} applies evolutionary computing to optimize parameters in reinforcement learning. More specifically, evolutionary computing evaluates the fitness function of each set of parameters, to find out the optimal solution in parameter space iteratively. The fitness function in ECRL is defined as:
\begin{equation}
    fitness\_function = -2(average\_reward) + error,
\end{equation}
the formula combines two indexes which are average reward and error, to maximize the performance and stability of the reinforcement learning algorithm. 

\begin{figure*}[ht!]
  \centering
  \begin{minipage}{0.49\textwidth}
    \includegraphics[width=\linewidth]{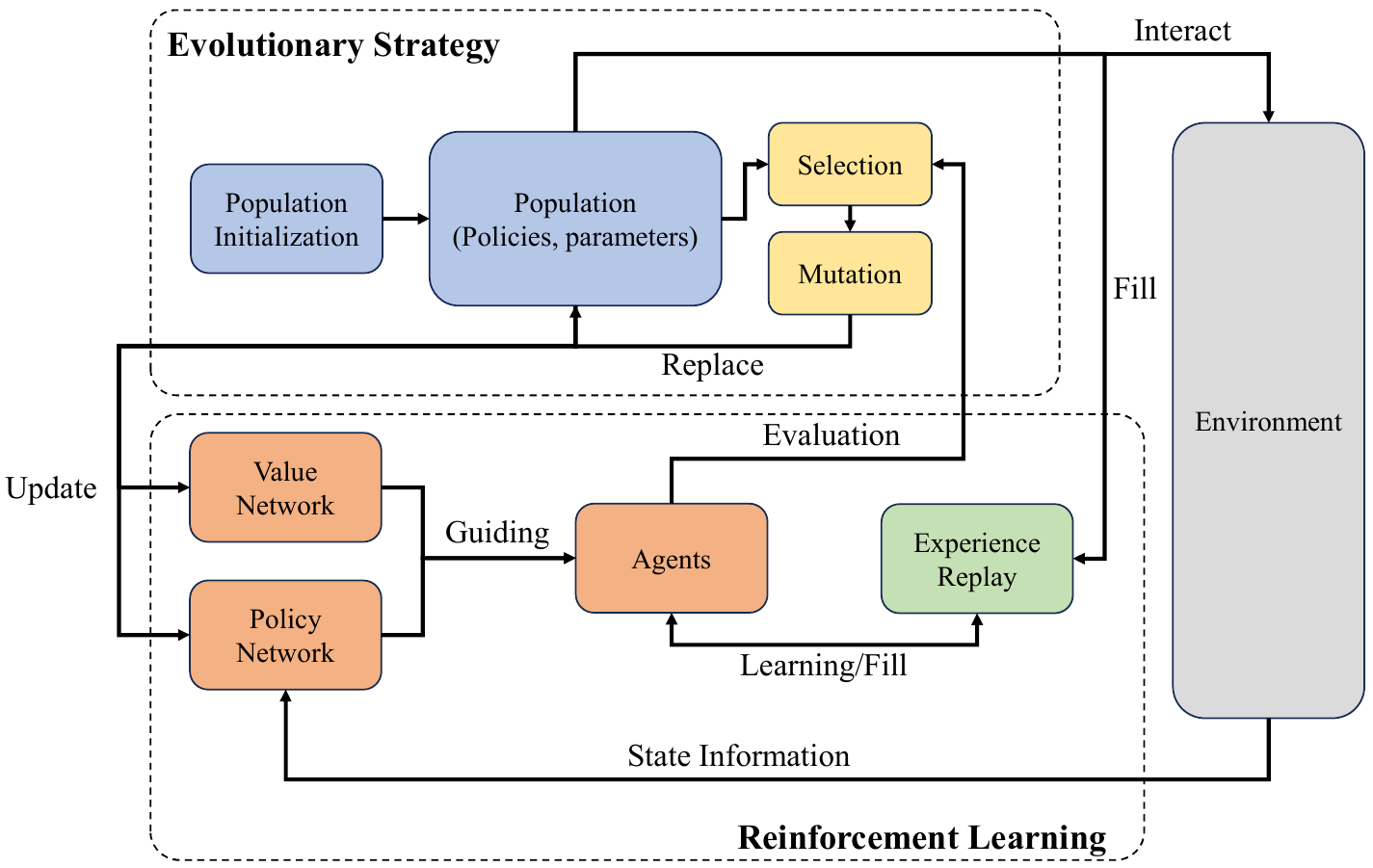} 
    \caption[A General Workflow of ES-based EvoRL]{A General Workflow of ES-based EvoRL.\\ The core evolutionary operation here is the mutation, \\which enhances the diversity of the population.}
    \label{ES}
  \end{minipage}\hfill
  \begin{minipage}{0.49\textwidth}
    \includegraphics[width=\linewidth]{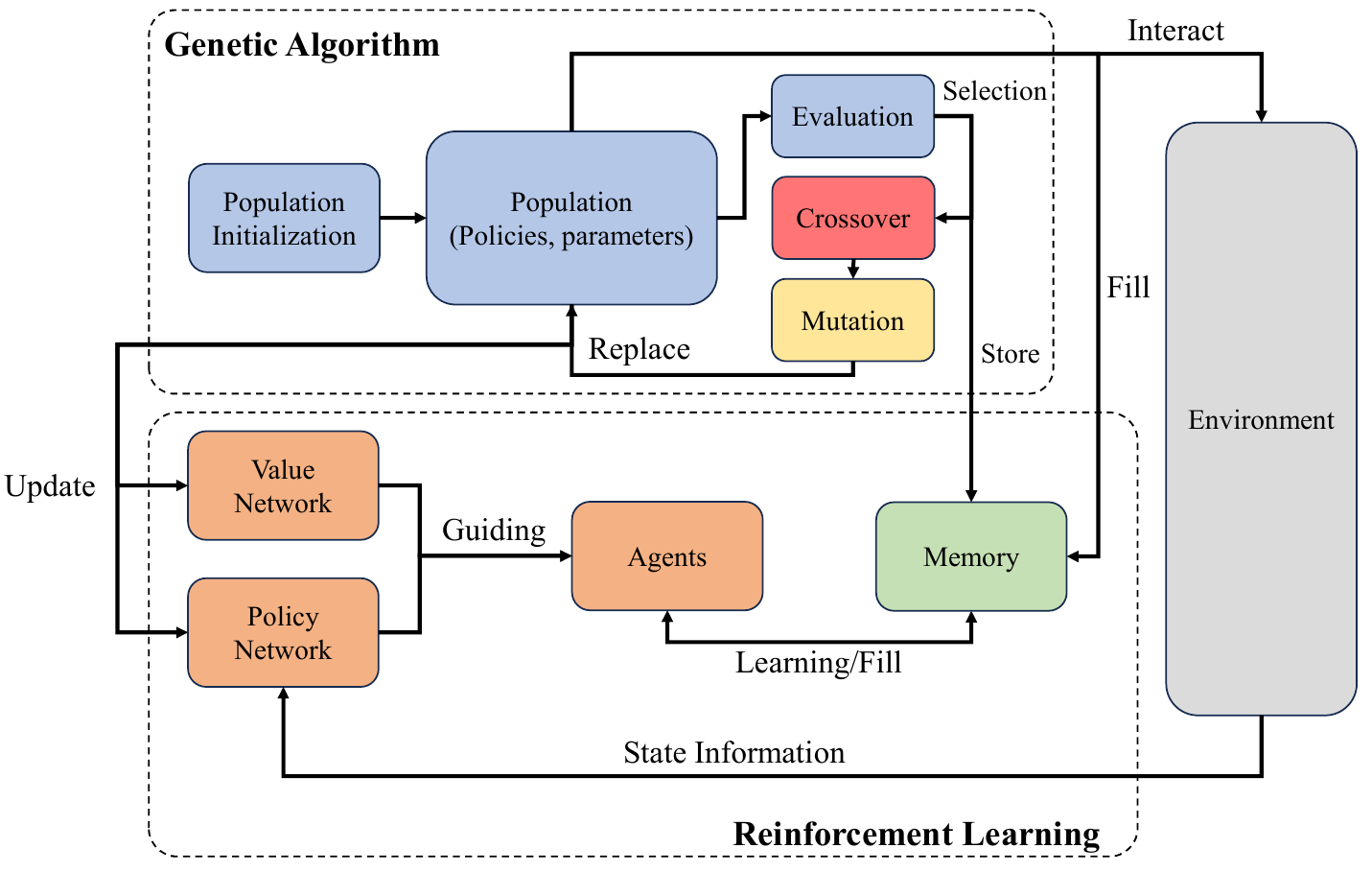} 
    \caption[A General Workflow of GA-based EvoRL]{A General Workflow of GA-based EvoRL.\\ The EvoRL retains good genes through crossover to \\produce better individuals.}
    \label{GA}
  \end{minipage}
\end{figure*}

We have discussed the core mechanism of EvoRL in the above contents, EvoRL applies EAs to optimize the decision process of reinforcement learning by simulating the operation of natural evolution. In this paper, evolutionary strategy, genetic algorithm, CEM and PBT primarily emphasize and explore their potential to develop effective and robust strategies in challenging environments. The general workflows of the most frequently used EvoRL methods, ES-based and GA-based, are depicted in Fig.\ref{ES} and Fig.\ref{GA}, respectively. Evolutionary strategy primarily emphasizes the direct optimization of policies and behaviors through the simulation of natural selection and genetic variation. In contrast, Genetic algorithm focuses on searching for effective and stable policies through genetic diversity and crossover operations. The CEM iteratively optimizes strategies through statistical methods, with a particular emphasis on elite selection to guide policy development. The PBT demonstrates flexibility in adjusting parameters and hyper-parameters, particularly in the automatic adjustment of reward functions. Random search improves exploration efficiency by simplifying the search process. In contrast, genetic programming exhibits a deeper complexity in the optimization of computation graphs. While these algorithms differ in methods and emphasis, they share a common core objective: to enhance the reinforcement learning process by applying principles from evolutionary theory, thereby improving the performance and adaptability of policies.
\section{Challenges in Evolutionary Reinforcement Learning}
In this chapter, we delve into the challenges encountered when employing reinforcement learning and EAs independently. We specifically analyze how these challenges underscore the importance and advantages of combining reinforcement learning and EA into EvoRL. While reinforcement learning and EA each exhibit significant strengths in addressing complex problems, they also have evident limitations. For instance, reinforcement learning commonly faces challenges such as parameter sensitivity, reward sparsity, susceptibility to local optima, multitask processing difficulties, policy search complexity, and computational efficiency issues. Similarly, when applied in isolation, EA encounters challenges such as hyper-parameter sensitivity, multi-objective optimization complexities, computational efficiency, and the design of fitness functions. To comprehensively understand these issues, this chapter is divided into two subsections, addressing the challenges faced when using reinforcement learning and EA individually. Furthermore, we explore how EvoRL, as a holistic approach, overcomes these limitations by integrating the decision optimization capabilities of reinforcement learning with the natural evolutionary simulation of EA, thereby achieving a more comprehensive, efficient, and adaptive problem-solving methodology. An overview of scientific problems encountered by both of algorithms and their corresponding resolutions is shown as Table \ref{tab:RLIssue} and Table \ref{tab:EvoIssue}.
\label{Challenges}
\begin{table*}
  \centering
  \caption{Scientific Issues Encountered by Reinforcement Learning and Corresponding EAs Solution}
  \footnotesize
    \resizebox{\linewidth}{!}{
    \begin{tabular}{lllll}
    \toprule
    \textbf{Scientific Issue} & \textbf{Resolution} & \textbf{Method} & \textbf{RL Algorithm} & \textbf{Evo Algorithm} \\
    \midrule
    \multirow{5}[3]{*}{Parameters Sensitivity} & \multirow{2}[0]{*}{EA dynamically adjusts parameters} & SBARL \cite{de2023out} & Policy gradient    & Population-based training \\
    & & MAEyS \cite{sachdeva2021maedys} & Policy gradient    & Evolutionary computing \\
    \cmidrule{2-5} 
          & \multirow{3}[0]{*}{EA enhances parameter space diversity}      & GA-DRL \cite{sehgal2019deep} & Actor-critic    & Genetic algorithms \\
          &         & CCNCS \cite{yang2022evolutionary}& Policy gradient    & Evolutionary strategy \\
        &       & NS-MERL \cite{aydeniz2023novelty} & Actor-critic    & Evolutionary strategy \\
    \midrule
    \multirow{4}[0]{*}{Sparse Reward} & \multirow{2}[0]{*}{EA accelerates the search process} & PS3-TD3 \cite{jung2020population} & Actor-critic    & Population-based training \\
          &       & PNS-RL \cite{liu2021pns} & Actor-critic    & Evolutionary strategy \\
\cmidrule{2-5}          &  \multirow{2}[0]{*}{EA enhances policy space diversity} & GEATL  \cite{zhu2021evolutionary} & Actor-critic    & Genetic algorithm \\
& & RACE \cite{li2023race} & Actor-cirtic & Genetic algorithm\\
    \midrule
    \multirow{3}[0]{*}{Local Optima} & \multirow{3}[0]{*}{EA enhances policy space diversity} & EARL \cite{wang2023evolutionary}  & Policy gradient    & Population-based training \\
          &  & G2AC \cite{chang2018genetic}  & Actor-critic    & Genetic algorithm \\
         &  & DEPRL \cite{liu2021diversity}    & Actor-critic    & Cross-entropy method \\
    \midrule
    \multirow{2}[0]{*}{Multi-task} & EA enhances parameter space diversity & A-MFEA-RL \cite{martinez2021adaptive} & Value function    & Genetic algorithm \\
\cmidrule{2-5}          & EA enhances policy combination and synergy & EGPRL \cite{kelly2021evolving} & Policy gradient    & Genetic programming \\
    \midrule
    \multirow{3}[0]{*}{Policy Search} & \multirow{3}[0]{*}{EA-integrated process for policy generation} & AGPRL \cite{kamio2005adaptation} & Value function    & Genetic programming \\
          &       & GPRL \cite{hein2018interpretable}  & Policy gradient    & Genetic programming \\
          &       & GPFDM \cite{girgin2008feature} & Policy gradient    & Genetic programming \\
    \midrule
    \multirow{3}[0]{*}{Computational Efficiency} & \multirow{3}[0]{*}{EA-integrated process for policy generation} & FiDi-RL \cite{shi2019fidi} & Actor-critic    & Random search \\
          &  & CERM-ACER \cite{tang2021guiding} & Actor-critic    & Cross-entropy method \\
         &  & CGP \cite{simmons2019q}  & Actor-critic    & Cross-entropy method \\
    \bottomrule
    \end{tabular}%
    }
  \label{tab:RLIssue}%
\end{table*}%

\begin{table*}[htbp]
  \centering
  \caption{Scientific Issues Encountered by Evolutionary Algorithms and Corresponding Reinforcement Learning Solution}
   \footnotesize
    \resizebox{\linewidth}{!}{
    \begin{tabular}{lllll}
    \toprule
    \textbf{Scientific Issue} & \textbf{Resolution} & \textbf{Method} & \textbf{RL Algorithm} & \textbf{Evo Algorithm} \\
    \midrule
    Hyper-parameters Sensitivity & RL guides the selection of evolutionary operators & ESAC \cite{suri2020maximum}  & Acor-critic & Evolutionary strategy \\
    \midrule
    \multirow{5}[3]{*}{Multi-objective} &\multirow{2}[0]{*}{RL guides the selection of evolutionary operators} & AMODE-DRL \cite{li2023scheduling} & Policy gradient & Genetic algorithm \\
    & & RL-MOEA \cite{zhang2023reinforcement} & Value function    & Genetic algorithm\\
\cmidrule{2-5}          & RL reduces the optimal solution search space & MOEADRL \cite{gao2023efficient} & Acor-critic & Genetic algorithm \\
\cmidrule{2-5}  & RL applies action to optimize the network& MERL \cite{yang2023reducing}  & Policy gradient    & Genetic algorithm \\
\cmidrule{2-5}
          & RL optimizes evolution process      & MOPDERL \cite{tran2023two}& Actor-critic    & Genetic algorithm \\
    \midrule
    \multirow{2}[0]{*}{Computational Efficiency} & RL applies action to optimize the network & BNEP \cite{stork2021behavior}  & Value function & Genetic programming \\
\cmidrule{2-5}          & RL guides the selection of evolutionary operators & RL-GA \cite{song2023rl} & Value function & Genetic algorithm \\
    \midrule
    \multirow{2}[0]{*}{Fitness Function} & RL applies the surrogate model to assist calculation & SMB-NE \cite{stork2019surrogate}& Policy gradient & Evolutionary strategy \\
    \cmidrule{2-5}          & RL optimizes evolution process & X-DDPG \cite{espositi2020gradient}      & Actor-critic    & Genetic algorithm \\
    \bottomrule
    \end{tabular}%
    }
  \label{tab:EvoIssue}%
\end{table*}%

\subsection{Issues Encountered by Reinforcement Learning}
When applying reinforcement learning independently, we face many of the scientific problems mentioned above. These challenges limit the scope and efficiency of reinforcement learning in complex environments, suggesting that further methodological innovations and technological advances are needed to overcome these limitations \cite{li2022reinforcement}. EA, as an algorithm that simulates biological evolutionary genetics, can help with reinforcement learning to some extent.

\subsubsection{Parameters Sensitivity}
The performance of reinforcement learning is highly related to the settings of its parameters including learning rate, discount factor, parameters of policy networks, parameters of reward function, etc. The adjustments and optimizations of these parameters are critical to the realization of an effective and stable reinforcement learning process. The unreasonable setting of parameters may lead to an unstable process of training, low speed of convergence, or not being able to find useful policy. EAs mainly dynamically adjust parameters in reinforcement learning through a series of evolutionary operations, or increase the diversity of parameter space to find the optimal combination of parameters.

According to \cite{de2023out}, the quality of final results in parameter control methods for metaheuristics with reinforcement learning is highly correlated with the values of these parameters. Therefore, \cite{de2023out} introduces the SBARL method, aiming to dynamically adjust and evolve these parameters while maintaining a static configuration. Specifically, the evaluation of PBT workers in SBARL aligns with their average reward in each training period. The less well-performing workers adopt the parameter settings of the better-performing workers as a reference for optimization and evolution. Similar to SBARL, \cite{sachdeva2021maedys} dynamically selects the optimal local skills to adapt to the varying requirements of a multi-agent environment, effectively addressing the parameter sensitivity issue in reinforcement learning. The proposed MAEDyS \cite{sachdeva2021maedys} not only utilizes policy gradient methods to learn multiple local skills but also enhances its capability to handle parameter sensitivity in complex multi-agent settings through the dynamic selection and optimization of these local skills.

Besides dynamic parameter adjustment, \cite{sehgal2019deep} employs a genetic algorithm to directly search and optimize parameters. The approach explores the parameter space through the selection, mutation, and crossover operations of the genetic algorithm, enhancing the diversity of the parameter space. This methodology aids in finding the optimal parameter combination for the proposed method. \cite{sehgal2019deep} presents Polyak-averaging coefficient updates as shown below:
\begin{equation}
    \begin{split}
        &\theta^{Q^\prime}\leftarrow \tau\theta^Q+(1-\tau)\theta^{Q^\prime}\\
        &\theta^{\mu^\prime}\leftarrow \tau\theta^\mu+(1-\tau)\theta^{\mu^\prime},
    \end{split}
\end{equation}
where $\theta^{Q^\prime}$ and $\theta^{\mu^\prime}$ are the parameters of the target Q-network and target policy network, respectively. $\theta^Q$ and $\theta^\mu$ denote the parameters of the current Q-network and policy network, $\tau$ is the Polyak-averaging coefficient, a factor usually close to but greater than zero, used to blend the current and target network parameters.

Likewise, \cite{yang2022evolutionary} focuses on optimizing parameters by enhancing the diversity of the parameter space. The approach employed by CCNCS \cite{yang2022evolutionary} utilizes a framework known as Cooperative Coevolution (CC) to break down large-scale optimization problems into smaller, more manageable sub-problems. CC enables the independent evolution of different parameter combinations while maintaining their interactions, facilitating the search for an optimal parameter solution.

\subsubsection{Sparse Reward}
The sparse reward problem is a significant challenge in reinforcement learning. As highlighted by \cite{liu2021pns}, agents face difficulty obtaining sufficient reward signals to guide an effective learning process when exploring environments with sparse rewards. In such scenarios, the decision-making process of agents may become inefficient due to the lack of immediate feedback, thereby impacting the performance and training speed \cite{liu2021pns}.

To address the issue, \cite{jung2020population} exploits the global searching ability of EA, particularly PBT, to expedite the search process through cooperation and information sharing among multi-agents. \cite{jung2020population} present the augmented loss function for an agent $i$ as: 
\begin{equation} 
    \Tilde{L}(\phi_i) = L(\phi_i) + 1_{{i=b}} \beta \mathbb{E}_{s \sim \mathcal{D}} D(\pi_{\phi_i}, \pi_{\phi_b}),
\end{equation} 
where $L(\phi_i)$ is the original loss function for agent $i$, $\beta$ is a constant, $E_{s \sim D}$ is the expectation over states sampled from distribution $\mathcal{D}$, $D(\pi_{\phi_i},\pi_{\phi_i})$ is a distance function measuring divergence between policies of agent $i$ and a baseline agent $b$, and $1_{\{ i=b\}}$ is an indicator function. More specifically, the proposed model accelerates the learning process by incorporating information on the best strategy, effectively guiding the agent to discover an effective strategy in sparse reward environments. 

In addition to expediting the search process, \cite{zhu2021evolutionary} introduces a GEATL method that employs a genetic algorithm to foster policy diversity through exploration in the parameter space. This policy exploration aids in uncovering solutions effective in sparse reward environments, as it doesn't depend on frequent or immediate reward feedback.

Similarly, \cite{li2023race} also resolve the sparse reward of multi-agent reinforcement learning through improving the diversity of policy space. \cite{li2023race} proposes a RACE method which considers each agent as a population, explores new policy space by the crossover and mutation of individuals in the population. The evolutionary operations are able to help reinforcement learning to generate diverse behavior modes. This kind of diversity is quite essential for those effective policy that only appears under certain conditions in sparse reward environment. \cite{li2023race} also introduces random perturbation, which is shown as:
\begin{equation}
    \begin{split}
        W^\prime_i,W^\prime_j&=((W_i-W_i^{d_i})\cup W_j^{d_i},(W_j-W_j^{d_j})\cup W_i^{d_j})\\
        &=\text{Crossover}(W_i,W_j),\\
        W^\prime_j&=(W_j-W_j^{d_j})\cup P(W_j^{d_j}=\text{Mutation}(W_j),\\
    \end{split}
\end{equation}
where $W_i$ and $W_j$ stand as the chosen teams, $di$ and $dj$ depict subsets of agent indices ${1, ..., N}$ selected at random. The perturbation function, denoted as $P$, introduces Gaussian noise to specific parameters or resets them. $W_d$ is utilized to denote the subset of policy representations corresponding to the team characterized by indices $d$.
\subsubsection{Local Optima}
The challenge of local optima is primarily attributed to the vanishing gradient during policy updating, as highlighted by \cite{liu2021diversity}. This issue may hinder effective exploration of better polices in complex environments. To handle this problem, \cite{liu2021diversity} introduces the DEPRL method, employing CEM to enhance policy diversity and improve exploration efficiency. This approach shows a significant improvement in continuous control tasks and effectively reduces the risk of getting trapped in local optima.

Similarly, \cite{wang2023evolutionary} introduces an approach that enhances policy diversity by integrating reinforcement learning and EAs within a unified framework. In the proposed EARL \cite{wang2023evolutionary}, the crucial concept lies in the exchange of information between reinforcement learning agents and EA populations. Reinforcement learning agents acquire diverse exploration experiences from the EA population, while the EA population regularly receives gradient information from reinforcement learning agents. This reciprocal interaction fosters strategy diversity, enhancing the stability and robustness of the algorithm. The formula of loss function in EARL is shown as:
\begin{equation}
    J_\pi(\phi)=\mathbb{E}_{\mathbf{s_t} \sim \mathcal{D}, \mathbf{a_t} \sim \pi_\phi(\cdot|\mathbf{s_t})} \lbrack \alpha \log \pi_\phi (\mathbf{a_t}|\mathbf{s_t})-\underset{j=1,2}{\min}Q_{\theta_j}(\mathbf{s_t},\mathbf{a_t})\rbrack,
\end{equation}
where $\mathbb{E}_{\mathbf{s_t} \sim \mathcal{D}, \mathbf{a_t} \sim \pi_\phi(\cdot|\mathbf{s_t})}$ represents the expectation over states $\mathbf{s_t}$ sampled from a dataset $\mathcal{D}$ and actions $\mathbf{a_t}$ sampled from the policy $\pi_\phi$ given the state $\mathbf{s_t}$.

For actor-critic algorithm, the diversity of policy space brought by EAs can improve sample efficiency and performance. To this end, \cite{chang2018genetic} proposes a G2AC approach that combines gradient-independent and gradient-dependent optimization by integrating genetic algorithms in hidden layers of neural networks. The policy update gradient of G2AC is given as following formula:
\begin{equation}
    \nabla_\theta J(\theta) = \mathbb{E_\pi}\lbrack\nabla_\theta \log \pi_\theta (s,a)A^\pi (s,a)\rbrack,
\end{equation}
where $\mathbb{E_\pi}$ denotes the expectation under the policy $\pi$ parameterized by $\theta$, $\nabla_\theta \log \pi_\theta (s,a)$ stands for the gradient of the logarithm of the policy $\pi_\theta$, evaluated at a specific state-action pair $(s,a)$,  $A^\pi (s,a)$ signifies the advantage function under policy $\pi$. This approach allows models to diversify their exploration in the solution space and jump around as they find better areas, and G2AC increases the diversity of policies in this way.

\subsubsection{Multi-task}
The multi-task challenges in reinforcement learning arise primarily from the dynamics and complexity of the real world, requiring agents to handle various tasks within distinct environments. As emphasized by \cite{kelly2021evolving}, agents may need to navigate environments with both discrete and continuous action spaces, often characterized by partial observability. To tackle this issue, \cite{kelly2021evolving} proposes the EGPRL method, combining multiple independently adapted agents to realize a synergistic effect among different policies, thereby enhancing multi-task performance. This is further supported by the Tangled Program Graph (TPG) framework in \cite{kelly2021evolving}, which leverages hierarchical structures and modular memory to effectively encode and manage environmental information in partially observable settings.

Similarly, \cite{martinez2021adaptive} introduces the A-MFEA-RL method. This approach employs a unified search space and genetic manipulation (e.g., crossover and mutation) to enhance the diversity of the parameter space, aiding reinforcement learning in exploring broader solutions in multi-task environments. Additionally, A-MFEA-RL utilizes adaptive knowledge transfer strategies to optimize the learning process across tasks, effectively balancing information exchange to prevent negative transfer while promoting beneficial synergies \cite{martinez2021adaptive}.

While the two mentioned articles propose distinct solutions to the multi-task challenge in reinforcement learning, they both employ EAs as a common strategy to enhance policies. Whether aiming to boost policy synergy or increase parameter space diversity for more effective policy search, both approaches demonstrate their efficacy in effectively addressing the multi-task issue.

\subsubsection{Policy Search}
In reinforcement learning, the policy search problem revolves around determining the optimal policy to maximize the cumulative reward for an agent interacting with its environment. A proficient policy search is pivotal for efficient learning and preventing potentially sub-optimal or erroneous behavior. As highlighted in \cite{hein2018interpretable}, conventional reinforcement learning approaches face challenges when confronted with intricate and high-dimensional state spaces. These difficulties can impede the learning process, resulting in subpar strategy quality.

Therefore, \cite{hein2018interpretable} proposes a GPRL method which is able to autonomously learn the policy equation. In this method, genetic programming is used to generate the basic algebraic equations that form the reinforcement lraning strategy from the pre-existing default state-action trajectory sample. The key to GPRL is that it learns interpretable and moderately complex policy representations from the data in the form of basic algebraic equations.

In a similar vein, genetic programming is applied in a two-stage process in another study \cite{kamio2005adaptation}. Initially, programs are generated in a simulated environment using genetic programming, serving as candidate solutions for various tasks. Subsequently, these actions, derived from genetic programming, are adapted to the operational characteristics of specific real robots through reinforcement learning, particularly Q-learning. The pivotal aspect of this approach is that the programs created by genetic programming provide an effective starting point for reinforcement learning, thereby accelerating and enhancing the process of policy search and adaptation.

Expanding on this concept, another study \cite{girgin2008feature} introduces a genetic programming-based method aimed at automating feature discovery in reinforcement learning. Central to this approach is the use of genetic programming to generate a set of features that significantly enhance the efficiency of reinforcement learning algorithms in learning strategies. The key lies in utilizing genetic programming to automatically unearth useful features from an agent's observations, capturing the intricate non-linear mappings between states and actions. This, in turn, improves both the efficiency and effectiveness of the policy search process \cite{girgin2008feature}.

\subsubsection{Computational Efficiency}
Traditional reinforcement learning methods often grapple with the high cost and inefficiency of calculating the derivative of the optimal target, leading to poor stability and robustness in complex tasks \cite{shi2019fidi}. This challenge is compounded when employing complex neural networks (NNs) as control strategies in most current approaches. These deep NNs, despite their potential for enhanced performance, complicate parameter tuning and computation. In response, \cite{shi2019fidi} introduces FiDi-RL, a novel method that integrates DRL with finite-difference (FiDi) policy search. By combining DDPG and Augmented Random Search (ARS), FiDi-RL enhances ARS's data efficiency. Empirical results validate that FiDi-RL not only boosts ARS's performance and stability but also stands competitively among existing DRL methods.

Complementing this, the CERM-ACER algorithm \cite{tang2021guiding} addresses computational efficiency in reinforcement learning through an EA perspective, blending the CEM with the actor-critic with experiential replay (ACER). This synergistic approach enables policy parameters to make substantial jumps in the parameter space, allowing for more assertive updates per iteration. Consequently, CERM-ACER not only stabilizes the algorithm but also diminishes the necessity for extensive sample collection in complex environments, thus boosting computational efficiency.

Similarly, the CGP algorithm \cite{simmons2019q} enhances the computational efficiency of Q-learning in continuous action domains. By fusing CEM with deterministic neural network strategies, CGP employs heuristic sampling for Q function training while simultaneously training a policy network to emulate CEM. This is mathematically represented by the L2 regression objective:
\begin{equation}
    J(\phi)=\mathbb{E}_{s\sim\rho^{\pi_{CEM}}}(\nabla_{\pi_\phi}\| \pi_\phi(s_t)-\pi_{CEM}(s_t) \|^2),
\end{equation}
where $J(\phi)$ denotes the objective function for training the policy network $\pi_\phi$, $s_t$ represents a state in the state space, $\pi_{CEM}$ is the policy generated by CEM, and $\rho^{\pi_{CEM}}$ is the distribution over states as determined by the CEM policy. 
This strategy eliminates the need for costly sample iterations during inference, significantly accelerating inference speed and reducing computational demands. CGP's efficacy in execution efficiency makes it particularly suited for real-time, compute-sensitive tasks.

In summary, these approaches demonstrate how EAs can revolutionize reinforcement learning's computational efficiency. Starting from policy generation, these methods adeptly navigate the complexities of reinforcement learning, offering more efficient, stable, and robust solutions.

\subsection{Issues Encountered by Evolutionary Algorithm}
Even though EAs has many advantages, its own shortcomings can not be ignored. Small changes of hyper-parameters may cause huge fluctuations in performance. Multi-objective optimization needs to strike a balance among conflicting objectives, and the limitation of computational efficiency is a problem that cannot be ignored. Designing a fitness function which can reflect the nature of the problem and is easy to calculate is one of the complex problems. The addition of reinforcement learning comes to the fore in these challenges, offering promising solutions to problems. Through adaptive learning, reinforcement learning assists EAs to find balance and optimize decision-making process in multi-objective and dynamic environment. Reinforcement learning shows high flexibility and efficiency in hyper-parameter adjustment and fitness function design. Therefore, EvoRL plays a key role in improving performance and application effectiveness.
\subsubsection{Hyper-parameters Sensitivity}
Hyper-parameter sensitivity is a key challenge where the algorithm's performance hinges on hyper-parameter settings. Even slight adjustments can cause notable variations in results, complicating tuning and limiting the algorithm's adaptability across tasks. Designers must invest significant time and resources in parameter tuning to find the best configuration \cite{suri2020maximum}.

Therefore, \cite{suri2020maximum} introduces the ESAC method, a combination of evolutionary strategy and SAC, to address hyper-parameter sensitivity. The reinforcement learning aspect in ESAC assists EA in overcoming this issue through its adaptive mechanism. Specifically, ESAC incorporates Automatic Mutation Tuning (AMT) to maximize the mutation rate, diminishing the need for precise hyper-parameter settings. The SmoothL1 loss function in AMT is defined as:
\begin{equation}
\begin{split}
Smooth L1(x_i,y_i)= \left \{
\begin{array}{ll}
    0.5(x_i-y_i)^2,                     & if |x_i-y_i| < 1\\
    |x_i-y_i|-0.5,     & otherwise\\
\end{array}
\right.
\end{split}
\end{equation}
The SmoothL1 loss function, a combination of absolute loss and square loss, aims to mitigate the impact of abnormal values on model training. Simultaneously, ESAC employs evolutionary strategies to explore the strategy in weight space and leverages the gradient-based knowledge of the SAC framework to utilize policies. This dual approach helps maintain a balance between exploration and exploitation, reducing reliance on a single hyper-parameter setting.

\subsubsection{Multi-objective}
Multi-objective optimization problem refers to how to effectively balance multiple conflicting objectives in an uncertain environment. For example, in some production environments \cite{zhang2023reinforcement}, the two goals of reducing costs and increasing production efficiency need to be considered simultaneously, and these goals often contradict each other, which increases the complexity of problem solving. The characteristic of this kind of multi-objective problem is that a way must be found to balance the interests of different objectives in order to achieve the optimal combined effect. This not only challenges algorithm designers, but also requires the effectiveness and feasibility of practical application scenarios.

Reinforcement learning offers a multi-faceted solution to address multi-objective problems in Evolutionary Algorithms (EAs), as outlined by \cite{zhang2023reinforcement}. The proposed RL-MOEA, built upon the NSGA-II framework, introduces reinforcement learning capabilities. The algorithm's core mechanism dynamically selects the most suitable crossover and mutation operators in each iteration, employing a strategy based on Q-learning. The central formula of the Q-learning strategy is:
\begin{equation}
    newQ_{s,a} = (1-\alpha)\cdot Q_{s,a} + \alpha \cdot (R_{s,a}+\gamma \cdot \max \Tilde{Q}_{s,a}),
\end{equation}
where $newQ_{s,a}$ and $Q_{s,a}$ represent the Q value after and before updating under state $s$ and action $a$. $\max \Tilde{Q}{s,a}$ denotes the maximal Q value of the next state, and $R{s,a}$ stands for the reward of taking action $a$ under state $s$. In multi-objective optimization, different crossover and mutation operators have varied effects on different targets. RL -MOEA utilizes Q-learning strategies to choose the most appropriate operator in each iteration, effectively balancing and optimizing individual objectives based on historical and current evaluation information.

In the realm of reinforcement aiding EAs in selecting evolutionary operators, \cite{li2023scheduling} proposes an AMODE-DRL method. This method integrates DRL into Multi-Objective Differential Evolution (MODE). Through this integration, DRL guides MODE in selecting suitable mutation operators and parameters, resulting in improved solutions compared to other multi-objective evolution and adaptive MODE algorithms.

Building upon the concept of reinforcement learning aiding evolutionary algorithms in selecting evolutionary operators, as exemplified by \cite{li2023scheduling} with their AMODE-DRL method, \cite{gao2023efficient} take a different approach. In their study, the integration of the actor-critic from DRL into evolutionary algorithms specifically targets the challenge of multi-objective optimization. By strategically focusing on non-zero decision variables and employing a reward function that compares HyperVolume (HV) indicators between states, the actor-critic model effectively narrows down the search space. This method not only streamlines the optimization process but also ensures that the evolutionary algorithm is directed towards the most relevant and impactful solutions, enhancing both efficiency and outcome quality in multi-objective scenarios.

Different from optimizing solution search space, \cite{yang2023reducing} utilizes reinforcement learning to refine neural network policy parameters. This approach targets a dual-objective issue in financial cloud services, focusing on load imbalance and server idle time. Their reinforcement learning model, involving routing decisions across servers, eschews the need for intermediate rewards by employing cumulative objectives as the reward metric. This strategy effectively narrows down the optimal solution search space, enhancing efficiency in tackling multi-objective optimization challenges in EAs.

Reinforcement learning can also address the multi-objective issue through optimizing the evolution process. In \cite{tran2023two}, the problem of multi-objective optimization can be defined as:
\begin{equation}
    \underset{\pi}{\max} F(\pi)=\underset{\pi}{\max} \lbrack f_1(\pi),f_2(\pi),...,f_m(\pi)\rbrack,
\end{equation}
where $m$ denotes the number of objectives, $\pi$ represents the policy, $f_i(\pi)=J^\pi_i$. The study uses a two-stage framework in evolutionary algorithms, where reinforcement learning optimizes both network parameters and evolutionary operator selection. This approach reduces the search space in multi-objective optimization. Crucially, the NSGA-II algorithm is employed for selection and sorting, enhancing solution diversity and quality. 

\subsubsection{Computational Efficiency}
In \cite{stork2019surrogate}, a significant challenge in EAs is their computational inefficiency, particularly when the problem requires a large number of parallel evaluations and is fast to evaluate. To address this, the study integrates reinforcement learning with gene programming. The advantage calculation in these methods is given by:
\begin{equation}
    A_\phi(s_t,a_t)=R_t(s_t,a_t)-V_\phi(s_t),
\end{equation}
where $R_t(s_t,a_t)$ represents the reward function, $V_\phi(s_t),$ is the state value estimated by the critic. This targeted approach enhances the efficiency of finding optimal solutions in EAs, particularly in complex environments with large search spaces.

In \cite{song2023rl}, the study transitions from optimizing networks in reinforcement learning to guiding the selection of evolutionary operators in EAs to enhance computational efficiency. They adopt a Q-learning method in reinforcement learning to direct the evolutionary process. This approach addresses the computational inefficiency of EAs, particularly evident when multiple parallel evaluations are required. By utilizing Q-learning, the study effectively selects mutation operators for optimization, where the specific operation for each individual's evolution is determined based on Q-values. This method ensures a more focused and efficient search within the evolutionary process.

\subsubsection{Fitness Function}
\cite{stork2019surrogate} points out that one disadvantage of EAs is that it can require a lot of functional evaluation and does not fully utilize the information available in each fitness evaluation, especially when fitness evaluation is costly. Therefore, \cite{stork2019surrogate} proposes a SMB-NE method, which applies reinforcement learning to deal with this issue. The proposed method is to partially replace the expensive fitness function by using a Surrogate Model-Based Optimization (SMBO). These proxy models utilize data-driven models to simulate fitness functions, thereby reducing evaluation costs. In MoutainCar mission mentioned in \cite{stork2019surrogate}, the key formula is:
\begin{equation}
    fitness:y(x)=-({maxHeight}_{episode} + \frac{{Reward}_{episode}}{100}),
\end{equation}
it signifies the altered fitness function, where $y(x)$ denotes the output of neural networks, ${maxHeight}_{episode}$ represents the maximal height reached in one attempt, ${Reward}_{episode}$ stands for the sum of reward obtained from this attempt.

Another article also pointed out the fitness function problem in EAs, especially genetic algorithm. \cite{espositi2020gradient} considers traditional genetic algorithm cannot effectively solve machine learning problems when the fitness function has a high variance. This results in solutions generated by genetic algorithm that are often less robust and perform poorly, which is known as the "generalization limitation" of genetic algorithm. To address the issue, \cite{espositi2020gradient} introduces a X-DDPG algorithm, which combines genetic algorithm and DDPG. At the heart of this hybrid approach is the effect of "gradient bias," which tilts the evolutionary process in favor of more robust solutions by periodically injecting agents trained with DDPG into genetic populations.
\section{Open Issues and Future Directions}
After a thorough review of EvoRL algorithms, it is evident that their current application does not stand out as a remarkable achievement but rather necessitates further refinement. In this section, we put forward some emerging topics for consideration.

\subsection{Open Issues}
\subsubsection{Scalability to High-Dimensional Spaces}
The challenge lies in extending EvoRL methodologies to effectively handle complex, high-dimensional action and state spaces commonly encountered in real-world applications such as autonomous vehicles \cite{rasouli2019autonomous}, Unmanned aerial vehicle \cite{bai2023towards}, and large-scale industrial systems. Overcoming this hurdle entails the development of EvoRL algorithms capable of efficiently exploring and exploiting these expansive spaces while maintaining computational tractability. Furthermore, ensuring the scalability of EvoRL necessitates the implementation of innovative techniques to handle the curse of dimensionality, facilitate effective knowledge transfer \cite{li2021meta} across related tasks, and enable the discovery of meaningful solutions amidst the inherent complexity of high-dimensional environments.

\subsubsection{Adaptability to Dynamic Environments}
Adaptability to dynamic environments stands out as a significant open issue in EvoRL. EvoRL systems usually face challenges in rapidly adjusting their policies to keep pace with changes in the environment, where the optimal strategy may evolve over time. As real-world applications often involve dynamic and uncertain conditions, resolving the challenge of adaptability is essential for making EvoRL systems robust and versatile in handling the complexities of changing environments. To this end, it requires the development of algorithms that can dynamically adapt to shifting conditions. Evolutionary algorithms with dynamic parameter adaptation \cite{aleti2016systematic}, such as Adaptive Evolution Strategies \cite{zhan2019adaptive}, represent one avenue of exploration. These methods allow the algorithm to autonomously adjust parameters based on environmental changes. Additionally, research might delve into the integration of memory mechanisms (\eg Long Short-Term Memory network \cite{hochreiter1997long}) or continual learning approaches to retain information from past experiences, enabling EvoRL agents to adapt more effectively to evolving scenarios.

\subsubsection{Sample Efficiency and Data Efficiency}
A notable open issue in EvoRL pertains to sample efficiency and data efficiency, where EvoRL algorithms often require a substantial number of interactions with the environment to learn effective policies. Addressing this challenge involves exploring innovative algorithms to enhance the efficiency of learning processes. One feasible solution is data expansion. For example, the dual training \cite{shi2022reinforcement} can be adopted to get different behavior distribution by updating the trajectory generator. Thus, the sampled trajectories are repeatedly used to optimize the policy. Moreover, the exploration of transfer learning techniques \cite{weiss2016survey} might improve the ability of EvoRL agents to leverage knowledge gained from previous tasks, thus reducing the demand for extensive data collection. The ongoing pursuit of algorithms that strike a balance between exploration and exploitation, combined with approaches emphasizing effective knowledge transfer, remains pivotal in advancing the sample and data efficiency of EvoRL to make it more practical and applicable to real-world scenarios.

\subsubsection{Adversarial Robustness in EvoRL}
How to ensure that EvoRL agents maintain resilience in the face of intentional perturbations or adversarial interventions, is another open issue in EvoRL. Unlike traditional adversarial attacks in DRL, the unique characteristics of EvoRL algorithms introduce a set of challenges that demand tailored solutions \cite{ajao2023secure}. Addressing this issue involves developing algorithms that can evolve policies capable of withstanding adversarial manipulations, ultimately leading to more reliable and secure decision in dynamic and uncertain environments. To this end, we may focus on training EvoRL agents with diverse adversarial examples, which promote transferable defenses that can withstand perturbations across different environments. Besides, it is worth designing evolutionary algorithms that emphasize safe exploration, aiming to guide the learning process towards policies that are less prone to adversarial manipulation.

\subsubsection{Ethic and Fairness}
Another open issue in EvoRL that demands attention is the ethic and fairness of evolved policies. As EvoRL applications become more pervasive, ensuring that the learned policies align with ethical standards and exhibit fairness is crucial. Ethical concerns may arise if evolved agents exhibit biased behavior or inadvertently learn strategies that have undesirable societal implications. To address this issue, researchers need to explore algorithms that incorporate fairness-aware objectives during the evolutionary process. Techniques inspired by fairness-aware machine learning, such as federated adversarial debiasing \cite{hong2021federated} or reweighted optimization \cite{petrovic2022fair}, could be adapted to the EvoRL context. Additionally, integrating human-in-the-loop approaches to validate and guide the evolutionary process may contribute to more ethically aligned policies. As EvoRL continues to impact diverse domains, it becomes imperative to develop algorithms that not only optimize for performance but also adhere to ethical considerations and ensure fairness in decision-making processes.

\subsection{Future Directions}
\subsubsection{Meta-Evolutionary Strategies}
Meta-evolutionary strategies involve the evolution of the parameters guiding the evolutionary process or even the evolution of entire learning algorithms. This approach enables EvoRL agents to adapt their behaviors across different tasks and environments, making it inherently more versatile. Techniques inspired by meta-learning, such as Model-Agnostic Meta-Learning algorithm \cite{finn2017model} applied to evolutionary algorithms, hold promise for enhancing the ability of agents to generalize knowledge across various tasks. By evolving strategies that can rapidly adapt to new challenges, the future of EvoRL lies in creating agents that not only excel in specific environments but can dynamically adjust and learn efficiently in diverse and evolving scenarios.

\subsubsection{Self-Adaptation and Self-Improvement Mechanisms}
In the future, EvoRL is likely to witness significant progress in the incorporation of self-adaptation and self-improvement mechanisms, reflecting a paradigm shift towards more autonomous and adaptive learning systems. Researchers are exploring algorithms that enable EvoRL agents to dynamically adjust their strategies and parameters without external intervention. Evolutionary algorithms with self-adaptation mechanisms, such as Self-Adaptive Differential Evolution \cite{elsayed2012improved} or hybrid differential evolution based on adaptive Q-Learning \cite{peng2023reinforcement}, exemplify this trend. These algorithms allow the optimization process to autonomously adapt to the characteristics of the problem at hand, enhancing efficiency and robustness. Additionally, the integration of self-improvement mechanisms, inspired by principles of continual learning, may empower EvoRL agents to accumulate knowledge over time and refine their policies iteratively. As self-adaptive and self-improving algorithms become integral to the EvoRL landscape, the future holds the promise of more resilient, efficient, and increasingly autonomous learning systems capable of thriving in complex and dynamic environments.

\subsubsection{Transfer Learning and Generalization}
It is poised to witness significant strides in the realms of transfer learning and generalization, essential for adapting EvoRL agents to a broader range of tasks and environments. Techniques inspired by transfer learning \cite{niu2020decade}, such as methods leveraging meta-learning for adaptation \cite{lian2019towards}, are likely to be at the forefront of this development. The goal is to equip EvoRL agents with the ability to generalize learned knowledge and skills, enabling them to solve tasks more effectively and rapidly in novel scenarios. As real-world applications demand greater flexibility and adaptability, the integration of transfer learning and generalization mechanisms will be pivotal in establishing EvoRL as a robust and versatile learning paradigm capable of addressing diverse challenges.

\subsubsection{Heterogeneous Networks and Multi-agent Systems}
As we look ahead, one key area of development involves extending EvoRL methodologies to address the complexities of diverse, heterogeneous environments where agents exhibit varying capabilities, goals, and behaviors. Embracing this heterogeneity requires evolving EvoRL algorithms that can adapt to different agent types, preferences, and constraints, thus enabling the emergence of more robust and adaptive collective behavior \cite{zhang2023coordinated}. Additionally, the advancement of EvoRL in multi-agent systems will entail exploring algorithms capable of learning effective coordination and cooperation strategies among diverse agents \cite{yang2023evolutionary}, fostering the evolution of sophisticated group behaviors while considering emergent properties and system-level objectives. This evolution in EvoRL will likely contribute significantly to addressing real-world challenges across domains such as autonomous systems, smart cities, and decentralized networks, paving the way for more resilient, scalable, and adaptable multi-agent ecosystems.

\subsubsection{Interpretability and Explainability}
The future trajectory of EvoRL is poised to place a heightened emphasis on interpretability and explainability, acknowledging the growing importance of transparent decision-making in artificial intelligence systems. One potential avenue involves the incorporation of symbolic reasoning~\cite{malandri2023model} during the evolutionary process, facilitating the generation of policies that are not only effective but also comprehensible to humans. Besides, hybrid approaches, merging EvoRL with rule-based methods~\cite{porebski2022evaluation}, may offer a synergistic solution, ensuring the emergence of policies that align with domain-specific knowledge and are more readily understandable. The integration of explainable meta-learning techniques~\cite{shao2023effect} could also play a role, enabling EvoRL agents to adapt their strategies to different tasks while maintaining trustworthy.

\subsubsection{Incorporating Large Language Models}
Incorporating EvoRL within large language models~\cite{min2021recent}, such as GPT-4~\cite{chang2023examining}, holds tremendous potential. For instance, we can leverage EvoRL to facilitate the evolution and adaptation of language model architectures that can effectively comprehend, generate, and respond to human language. By integrating EvoRL with large language models, we can anticipate advancements in training methods that enable these models to not only learn from explicit rewards but also understand implicit signals within human interactions. This fusion could lead to the development of language models with enhanced contextual understanding, ethical reasoning capabilities, and the capacity to engage in more meaningful and empathetic interactions with users. Furthermore, the synergy between EvoRL and large language models may pave the way for novel applications in conversational artificial intelligence, personalized content generation, and context-aware decision-making systems. For instance, we can extend EvoRL to incorporate multi-modal information from large language models~\cite{zhong2023adapter}, aiming to shape more intuitive and socially aware artificial intelligence interfaces.
\section{Conclusion}
\label{Conclusion}
As an emerging technology, EvoRL holds considerable promise across diverse tasks. In light of this, our paper presents a comprehensive review of various EvoRL algorithms. Initially, we provide a background on EAs and reinforcement learning, serving as the foundation for EvoRLs. Subsequently, we categorize existing EvoRLs, highlighting the corresponding EAs they employ. Within each subdomain, we organize them based on different reinforcement learning methods, establishing a systematic classification rule for EvoRL. Furthermore, we explore challenges faced by EAs and reinforcement learning independently. Within this context, we examine potential solutions proposed by EvoRLs. To a certain extent, these challenges impede the performance of EAs and reinforcement learning, and EvoRLs offer related solutions that exhibit superior adaptability and problem-solving capabilities under specific circumstances. Finally, we put forward several emerging topics, including constructive open issues and promising future directions for the development of EvoRL, aiming to enhance its performance across an expanding array of scenarios.





\newpage


\vspace{11pt}


\bibliographystyle{IEEEtran}
\bibliography{IEEEabrv,reference}

\begin{thebibliography}{100}
\providecommand{\url}[1]{#1}
\csname url@samestyle\endcsname
\providecommand{\newblock}{\relax}
\providecommand{\bibinfo}[2]{#2}
\providecommand{\BIBentrySTDinterwordspacing}{\spaceskip=0pt\relax}
\providecommand{\BIBentryALTinterwordstretchfactor}{4}
\providecommand{\BIBentryALTinterwordspacing}{\spaceskip=\fontdimen2\font plus
\BIBentryALTinterwordstretchfactor\fontdimen3\font minus \fontdimen4\font\relax}
\providecommand{\BIBforeignlanguage}[2]{{%
\expandafter\ifx\csname l@#1\endcsname\relax
\typeout{** WARNING: IEEEtran.bst: No hyphenation pattern has been}%
\typeout{** loaded for the language `#1'. Using the pattern for}%
\typeout{** the default language instead.}%
\else
\language=\csname l@#1\endcsname
\fi
#2}}
\providecommand{\BIBdecl}{\relax}
\BIBdecl

\bibitem{kaelbling1996reinforcement}
L.~P. Kaelbling, M.~L. Littman, and A.~W. Moore, ``Reinforcement learning: A survey,'' \emph{Journal of artificial intelligence research}, vol.~4, pp. 237--285, 1996.

\bibitem{morimoto2005robust}
J.~Morimoto and K.~Doya, ``Robust reinforcement learning,'' \emph{Neural computation}, vol.~17, no.~2, pp. 335--359, 2005.

\bibitem{zhao2018deep}
X.~Zhao, L.~Xia, L.~Zhang, Z.~Ding, D.~Yin, and J.~Tang, ``Deep reinforcement learning for page-wise recommendations,'' in \emph{Proceedings of the 12th ACM conference on recommender systems}, 2018, pp. 95--103.

\bibitem{de2023out}
M.~G.~P. de~LACERDA, F.~B. de~Lima~Neto, T.~B. Ludermir, and H.~Kuchen, ``Out-of-the-box parameter control for evolutionary and swarm-based algorithms with distributed reinforcement learning,'' \emph{Swarm Intelligence}, pp. 1--45, 2023.

\bibitem{liu2021pns}
Q.~Liu, Y.~Wang, and X.~Liu, ``Pns: Population-guided novelty search for reinforcement learning in hard exploration environments,'' in \emph{2021 IEEE/RSJ International Conference on Intelligent Robots and Systems (IROS)}.\hskip 1em plus 0.5em minus 0.4em\relax IEEE, 2021, pp. 5627--5634.

\bibitem{goldberg1989cenetic}
D.~E. Goldberg, ``Cenetic algorithms in search,'' \emph{Optimization, Machine Learning}, 1989.

\bibitem{coello2007evolutionary}
C.~A.~C. Coello, \emph{Evolutionary algorithms for solving multi-objective problems}.\hskip 1em plus 0.5em minus 0.4em\relax Springer, 2007.

\bibitem{suri2020maximum}
K.~Suri, X.~Q. Shi, K.~N. Plataniotis, and Y.~A. Lawryshyn, ``Maximum mutation reinforcement learning for scalable control,'' \emph{arXiv preprint arXiv:2007.13690}, 2020.

\bibitem{zhang2023reinforcement}
Z.~Zhang, Q.~Tang, M.~Chica, and Z.~Li, ``Reinforcement learning-based multiobjective evolutionary algorithm for mixed-model multimanned assembly line balancing under uncertain demand,'' \emph{IEEE Transactions on Cybernetics}, 2023.

\bibitem{nilsson2021policy}
O.~Nilsson and A.~Cully, ``Policy gradient assisted map-elites,'' in \emph{Proceedings of the Genetic and Evolutionary Computation Conference}, 2021, pp. 866--875.

\bibitem{khadka2018evolution}
S.~Khadka and K.~Tumer, ``Evolution-guided policy gradient in reinforcement learning,'' \emph{Advances in Neural Information Processing Systems}, vol.~31, 2018.

\bibitem{shi2020efficient}
L.~Shi, S.~Li, Q.~Zheng, M.~Yao, and G.~Pan, ``Efficient novelty search through deep reinforcement learning,'' \emph{IEEE Access}, vol.~8, pp. 128\,809--128\,818, 2020.

\bibitem{khadka2019collaborative}
S.~Khadka, S.~Majumdar, T.~Nassar, Z.~Dwiel, E.~Tumer, S.~Miret, Y.~Liu, and K.~Tumer, ``Collaborative evolutionary reinforcement learning,'' in \emph{International conference on machine learning}.\hskip 1em plus 0.5em minus 0.4em\relax PMLR, 2019, pp. 3341--3350.

\bibitem{lu2021recruitment}
S.~L{\"u}, S.~Han, W.~Zhou, and J.~Zhang, ``Recruitment-imitation mechanism for evolutionary reinforcement learning,'' \emph{Information Sciences}, vol. 553, pp. 172--188, 2021.

\bibitem{franke2020sample}
J.~K. Franke, G.~K{\"o}hler, A.~Biedenkapp, and F.~Hutter, ``Sample-efficient automated deep reinforcement learning,'' \emph{arXiv preprint arXiv:2009.01555}, 2020.

\bibitem{gupta2021embodied}
A.~Gupta, S.~Savarese, S.~Ganguli, and L.~Fei-Fei, ``Embodied intelligence via learning and evolution,'' \emph{Nature communications}, vol.~12, no.~1, p. 5721, 2021.

\bibitem{pierrot2020sample}
T.~Pierrot, V.~Mac{\'e}, G.~Cideron, N.~Perrin, K.~Beguir, and O.~Sigaud, ``Sample efficient quality diversity for neural continuous control,'' 2020.

\bibitem{marchesini2020genetic}
E.~Marchesini, D.~Corsi, and A.~Farinelli, ``Genetic soft updates for policy evolution in deep reinforcement learning,'' in \emph{International Conference on Learning Representations}, 2020.

\bibitem{eriksson2003evolution}
A.~Eriksson, G.~Capi, and K.~Doya, ``Evolution of meta-parameters in reinforcement learning algorithm,'' in \emph{Proceedings 2003 IEEE/RSJ International Conference on Intelligent Robots and Systems (IROS 2003)(Cat. No. 03CH37453)}, vol.~1.\hskip 1em plus 0.5em minus 0.4em\relax IEEE, 2003, pp. 412--417.

\bibitem{tjanaka2022approximating}
B.~Tjanaka, M.~C. Fontaine, J.~Togelius, and S.~Nikolaidis, ``Approximating gradients for differentiable quality diversity in reinforcement learning,'' in \emph{Proceedings of the Genetic and Evolutionary Computation Conference}, 2022, pp. 1102--1111.

\bibitem{majid2023deep}
A.~Y. Majid, S.~Saaybi, V.~Francois-Lavet, R.~V. Prasad, and C.~Verhoeven, ``Deep reinforcement learning versus evolution strategies: a comparative survey,'' \emph{IEEE Transactions on Neural Networks and Learning Systems}, 2023.

\bibitem{wang2021evolutionary}
Y.~Wang, K.~Xue, and C.~Qian, ``Evolutionary diversity optimization with clustering-based selection for reinforcement learning,'' in \emph{International Conference on Learning Representations}, 2021.

\bibitem{sigaud2023combining}
O.~Sigaud, ``Combining evolution and deep reinforcement learning for policy search: a survey,'' \emph{ACM Transactions on Evolutionary Learning}, vol.~3, no.~3, pp. 1--20, 2023.

\bibitem{bai2023evolutionary}
H.~Bai, R.~Cheng, and Y.~Jin, ``Evolutionary reinforcement learning: A survey,'' \emph{Intelligent Computing}, vol.~2, p. 0025, 2023.

\bibitem{vikhar2016evolutionary}
P.~A. Vikhar, ``Evolutionary algorithms: A critical review and its future prospects,'' in \emph{2016 International conference on global trends in signal processing, information computing and communication (ICGTSPICC)}.\hskip 1em plus 0.5em minus 0.4em\relax IEEE, 2016, pp. 261--265.

\bibitem{hoffmeister1990genetic}
F.~Hoffmeister and T.~B{\"a}ck, ``Genetic algorithms and evolution strategies: Similarities and differences,'' in \emph{International conference on parallel problem solving from nature}.\hskip 1em plus 0.5em minus 0.4em\relax Springer, 1990, pp. 455--469.

\bibitem{vent1975rechenberg}
W.~Vent, ``Rechenberg, ingo, evolutionsstrategie—optimierung technischer systeme nach prinzipien der biologischen evolution. 170 s. mit 36 abb. frommann-holzboog-verlag. stuttgart 1973. broschiert,'' 1975.

\bibitem{zhu2023survey}
Q.~Zhu, X.~Wu, Q.~Lin, L.~Ma, J.~Li, Z.~Ming, and J.~Chen, ``A survey on evolutionary reinforcement learning algorithms,'' \emph{Neurocomputing}, vol. 556, p. 126628, 2023.

\bibitem{slowik2020evolutionary}
A.~Slowik and H.~Kwasnicka, ``Evolutionary algorithms and their applications to engineering problems,'' \emph{Neural Computing and Applications}, vol.~32, pp. 12\,363--12\,379, 2020.

\bibitem{ho2010cross}
S.~L. Ho and S.~Yang, ``The cross-entropy method and its application to inverse problems,'' \emph{IEEE transactions on magnetics}, vol.~46, no.~8, pp. 3401--3404, 2010.

\bibitem{botev2013cross}
Z.~I. Botev, D.~P. Kroese, R.~Y. Rubinstein, and P.~L’Ecuyer, ``The cross-entropy method for optimization,'' in \emph{Handbook of statistics}.\hskip 1em plus 0.5em minus 0.4em\relax Elsevier, 2013, vol.~31, pp. 35--59.

\bibitem{huang2021cem}
K.~Huang, S.~Lale, U.~Rosolia, Y.~Shi, and A.~Anandkumar, ``Cem-gd: Cross-entropy method with gradient descent planner for model-based reinforcement learning,'' \emph{arXiv preprint arXiv:2112.07746}, 2021.

\bibitem{jaderberg2017population}
M.~Jaderberg, V.~Dalibard, S.~Osindero, W.~M. Czarnecki, J.~Donahue, A.~Razavi, O.~Vinyals, T.~Green, I.~Dunning, K.~Simonyan \emph{et~al.}, ``Population based training of neural networks,'' \emph{arXiv preprint arXiv:1711.09846}, 2017.

\bibitem{hein2018interpretable}
D.~Hein, S.~Udluft, and T.~A. Runkler, ``Interpretable policies for reinforcement learning by genetic programming,'' \emph{Engineering Applications of Artificial Intelligence}, vol.~76, pp. 158--169, 2018.

\bibitem{sachdeva2021maedys}
E.~Sachdeva, S.~Khadka, S.~Majumdar, and K.~Tumer, ``Maedys: Multiagent evolution via dynamic skill selection,'' in \emph{Proceedings of the Genetic and Evolutionary Computation Conference}, 2021, pp. 163--171.

\bibitem{shi2019fidi}
L.~Shi, S.~Li, L.~Cao, L.~Yang, G.~Zheng, and G.~Pan, ``Fidi-rl: Incorporating deep reinforcement learning with finite-difference policy search for efficient learning of continuous control,'' \emph{arXiv preprint arXiv:1907.00526}, 2019.

\bibitem{lin2023comprehensive}
Y.~Lin, H.~Chen, W.~Xia, F.~Lin, P.~Wu, Z.~Wang, and Y.~Li, ``A comprehensive survey on deep learning techniques in educational data mining,'' \emph{arXiv preprint arXiv:2309.04761}, 2023.

\bibitem{chen2021survey}
X.~Chen, L.~Yao, J.~McAuley, G.~Zhou, and X.~Wang, ``A survey of deep reinforcement learning in recommender systems: A systematic review and future directions,'' \emph{arXiv preprint arXiv:2109.03540}, 2021.

\bibitem{lin2023survey}
Y.~Lin, Y.~Liu, F.~Lin, L.~Zou, P.~Wu, W.~Zeng, H.~Chen, and C.~Miao, ``A survey on reinforcement learning for recommender systems,'' \emph{IEEE Transactions on Neural Networks and Learning Systems}, 2023.

\bibitem{ajani2022adaptive}
O.~S. Ajani and R.~Mallipeddi, ``Adaptive evolution strategy with ensemble of mutations for reinforcement learning,'' \emph{Knowledge-Based Systems}, vol. 245, p. 108624, 2022.

\bibitem{tang2020online}
Y.~Tang and K.~Choromanski, ``Online hyper-parameter tuning in off-policy learning via evolutionary strategies,'' \emph{arXiv preprint arXiv:2006.07554}, 2020.

\bibitem{chen2019restart}
Z.~Chen, Y.~Zhou, X.~He, and S.~Jiang, ``A restart-based rank-1 evolution strategy for reinforcement learning.'' in \emph{IJCAI}, 2019, pp. 2130--2136.

\bibitem{yang2022evolutionary}
P.~Yang, H.~Zhang, Y.~Yu, M.~Li, and K.~Tang, ``Evolutionary reinforcement learning via cooperative coevolutionary negatively correlated search,'' \emph{Swarm and Evolutionary Computation}, vol.~68, p. 100974, 2022.

\bibitem{su2023evolution}
C.~Su, C.~Zhang, D.~Xia, B.~Han, C.~Wang, G.~Chen, and L.~Xie, ``Evolution strategies-based optimized graph reinforcement learning for solving dynamic job shop scheduling problem,'' \emph{Applied Soft Computing}, vol. 145, p. 110596, 2023.

\bibitem{sun2020zeroth}
H.~Sun, Z.~Xu, Y.~Song, M.~Fang, J.~Xiong, B.~Dai, and B.~Zhou, ``Zeroth-order supervised policy improvement,'' \emph{arXiv preprint arXiv:2006.06600}, 2020.

\bibitem{houthooft2018evolved}
R.~Houthooft, Y.~Chen, P.~Isola, B.~Stadie, F.~Wolski, O.~Jonathan~Ho, and P.~Abbeel, ``Evolved policy gradients,'' \emph{Advances in Neural Information Processing Systems}, vol.~31, 2018.

\bibitem{callaghan2023evolutionary}
A.~Callaghan, K.~Mason, and P.~Mannion, ``Evolutionary strategy guided reinforcement learning via multibuffer communication,'' \emph{arXiv preprint arXiv:2306.11535}, 2023.

\bibitem{martinez2021adaptive}
A.~D. Martinez, J.~Del~Ser, E.~Osaba, and F.~Herrera, ``Adaptive multifactorial evolutionary optimization for multitask reinforcement learning,'' \emph{IEEE Transactions on Evolutionary Computation}, vol.~26, no.~2, pp. 233--247, 2021.

\bibitem{li2023multi}
W.~Li, S.~He, X.~Mao, B.~Li, C.~Qiu, J.~Yu, F.~Peng, and X.~Tan, ``Multi-agent evolution reinforcement learning method for machining parameters optimization based on bootstrap aggregating graph attention network simulated environment,'' \emph{Journal of Manufacturing Systems}, vol.~67, pp. 424--438, 2023.

\bibitem{zheng2023rethinking}
B.~Zheng and R.~Cheng, ``Rethinking population-assisted off-policy reinforcement learning,'' \emph{arXiv preprint arXiv:2305.02949}, 2023.

\bibitem{elfwing2008co}
S.~Elfwing, E.~Uchibe, K.~Doya, and H.~I. Christensen, ``Co-evolution of shaping rewards and meta-parameters in reinforcement learning,'' \emph{Adaptive Behavior}, vol.~16, no.~6, pp. 400--412, 2008.

\bibitem{li2023scheduling}
T.~Li, Y.~Meng, and L.~Tang, ``Scheduling of continuous annealing with a multi-objective differential evolution algorithm based on deep reinforcement learning,'' \emph{IEEE Transactions on Automation Science and Engineering}, 2023.

\bibitem{yang2023reducing}
P.~Yang, L.~Zhang, H.~Liu, and G.~Li, ``Reducing idleness in financial cloud via multi-objective evolutionary reinforcement learning based load balancer,'' \emph{arXiv preprint arXiv:2305.03463}, 2023.

\bibitem{garau2022multi}
J.~J. Garau-Luis, Y.~Miao, J.~D. Co-Reyes, A.~Parisi, J.~Tan, E.~Real, and A.~Faust, ``Multi-objective evolution for generalizable policy gradient algorithms,'' in \emph{ICLR 2022 Workshop on Generalizable Policy Learning in Physical World}, 2022.

\bibitem{li2023evolutionary}
J.~Li and T.~Zhou, ``Evolutionary multi agent deep meta reinforcement learning method for swarm intelligence energy management of isolated multi area microgrid with internet of things,'' \emph{IEEE Internet of Things Journal}, 2023.

\bibitem{hu2023evolving}
C.~Hu, J.~Pei, J.~Liu, and X.~Yao, ``Evolving constrained reinforcement learning policy,'' \emph{arXiv preprint arXiv:2304.09869}, 2023.

\bibitem{bodnar2020proximal}
C.~Bodnar, B.~Day, and P.~Li{\'o}, ``Proximal distilled evolutionary reinforcement learning,'' in \emph{Proceedings of the AAAI Conference on Artificial Intelligence}, vol.~34, no.~04, 2020, pp. 3283--3290.

\bibitem{wang2022surrogate}
Y.~Wang, T.~Zhang, Y.~Chang, X.~Wang, B.~Liang, and B.~Yuan, ``A surrogate-assisted controller for expensive evolutionary reinforcement learning,'' \emph{Information Sciences}, vol. 616, pp. 539--557, 2022.

\bibitem{pierrot2022diversity}
T.~Pierrot, V.~Mac{\'e}, F.~Chalumeau, A.~Flajolet, G.~Cideron, K.~Beguir, A.~Cully, O.~Sigaud, and N.~Perrin-Gilbert, ``Diversity policy gradient for sample efficient quality-diversity optimization,'' in \emph{Proceedings of the Genetic and Evolutionary Computation Conference}, 2022, pp. 1075--1083.

\bibitem{hao2022erl}
J.~Hao, P.~Li, H.~Tang, Y.~Zheng, X.~Fu, and Z.~Meng, ``Erl-re$^2$: Efficient evolutionary reinforcement learning with shared state representation and individual policy representation,'' \emph{arXiv preprint arXiv:2210.17375}, 2022.

\bibitem{kalashnikov2018qt}
D.~Kalashnikov, A.~Irpan, P.~Pastor, J.~Ibarz, A.~Herzog, E.~Jang, D.~Quillen, E.~Holly, M.~Kalakrishnan, V.~Vanhoucke \emph{et~al.}, ``Qt-opt: Scalable deep reinforcement learning for vision-based robotic manipulation,'' \emph{arXiv preprint arXiv:1806.10293}, 2018.

\bibitem{shao2022grac}
L.~Shao, Y.~You, M.~Yan, S.~Yuan, Q.~Sun, and J.~Bohg, ``Grac: Self-guided and self-regularized actor-critic,'' in \emph{Conference on Robot Learning}.\hskip 1em plus 0.5em minus 0.4em\relax PMLR, 2022, pp. 267--276.

\bibitem{pourchot2018cem}
A.~Pourchot and O.~Sigaud, ``Cem-rl: Combining evolutionary and gradient-based methods for policy search,'' \emph{arXiv preprint arXiv:1810.01222}, 2018.

\bibitem{kim2020pgps}
N.~Kim, H.~Baek, and H.~Shin, ``Pgps: Coupling policy gradient with population-based search,'' 2020.

\bibitem{zheng2020cooperative}
H.~Zheng, P.~Wei, J.~Jiang, G.~Long, Q.~Lu, and C.~Zhang, ``Cooperative heterogeneous deep reinforcement learning,'' \emph{Advances in Neural Information Processing Systems}, vol.~33, pp. 17\,455--17\,465, 2020.

\bibitem{shi2021soft}
Z.~Shi and S.~P. Singh, ``Soft actor-critic with cross-entropy policy optimization,'' \emph{arXiv preprint arXiv:2112.11115}, 2021.

\bibitem{liu2021diversity}
J.~Liu and L.~Feng, ``Diversity evolutionary policy deep reinforcement learning,'' \emph{Computational Intelligence and Neuroscience}, vol. 2021, pp. 1--11, 2021.

\bibitem{wang2023evolutionary}
Z.-Z. Wang, K.~Zhang, G.-D. Chen, J.-D. Zhang, W.-D. Wang, H.-C. Wang, L.-M. Zhang, X.~Yan, and J.~Yao, ``Evolutionary-assisted reinforcement learning for reservoir real-time production optimization under uncertainty,'' \emph{Petroleum Science}, vol.~20, no.~1, pp. 261--276, 2023.

\bibitem{ma2022evolutionary}
Y.~Ma, T.~Liu, B.~Wei, Y.~Liu, K.~Xu, and W.~Li, ``Evolutionary action selection for gradient-based policy learning,'' in \emph{International Conference on Neural Information Processing}.\hskip 1em plus 0.5em minus 0.4em\relax Springer, 2022, pp. 579--590.

\bibitem{jung2020population}
W.~Jung, G.~Park, and Y.~Sung, ``Population-guided parallel policy search for reinforcement learning,'' \emph{arXiv preprint arXiv:2001.02907}, 2020.

\bibitem{doan2019attraction}
T.~Doan, B.~Mazoure, M.~Abdar, A.~Durand, J.~Pineau, and R.~D. Hjelm, ``Attraction-repulsion actor-critic for continuous control reinforcement learning,'' \emph{arXiv preprint arXiv:1909.07543}, 2019.

\bibitem{marchesini2022exploring}
E.~Marchesini, D.~Corsi, and A.~Farinelli, ``Exploring safer behaviors for deep reinforcement learning,'' in \emph{Proceedings of the AAAI Conference on Artificial Intelligence}, vol.~36, no.~7, 2022, pp. 7701--7709.

\bibitem{majumdar2020evolutionary}
S.~Majumdar, S.~Khadka, S.~Miret, S.~McAleer, and K.~Tumer, ``Evolutionary reinforcement learning for sample-efficient multiagent coordination,'' in \emph{International Conference on Machine Learning}.\hskip 1em plus 0.5em minus 0.4em\relax PMLR, 2020, pp. 6651--6660.

\bibitem{long2020evolutionary}
Q.~Long, Z.~Zhou, A.~Gupta, F.~Fang, Y.~Wu, and X.~Wang, ``Evolutionary population curriculum for scaling multi-agent reinforcement learning,'' \emph{arXiv preprint arXiv:2003.10423}, 2020.

\bibitem{shen2020generating}
R.~Shen, Y.~Zheng, J.~Hao, Z.~Meng, Y.~Chen, C.~Fan, and Y.~Liu, ``Generating behavior-diverse game ais with evolutionary multi-objective deep reinforcement learning.'' in \emph{IJCAI}, 2020, pp. 3371--3377.

\bibitem{fernandez2018parameters}
F.~C. Fernandez and W.~Caarls, ``Parameters tuning and optimization for reinforcement learning algorithms using evolutionary computing,'' in \emph{2018 International Conference on Information Systems and Computer Science (INCISCOS)}.\hskip 1em plus 0.5em minus 0.4em\relax IEEE, 2018, pp. 301--305.

\bibitem{kamio2005adaptation}
S.~Kamio and H.~Iba, ``Adaptation technique for integrating genetic programming and reinforcement learning for real robots,'' \emph{IEEE Transactions on Evolutionary Computation}, vol.~9, no.~3, pp. 318--333, 2005.

\bibitem{co2021evolving}
J.~D. Co-Reyes, Y.~Miao, D.~Peng, E.~Real, S.~Levine, Q.~V. Le, H.~Lee, and A.~Faust, ``Evolving reinforcement learning algorithms,'' \emph{arXiv preprint arXiv:2101.03958}, 2021.

\bibitem{abuzekry2019comparative}
A.~AbuZekry, I.~Sobh, M.~Hadhoud, and M.~Fayek, ``Comparative study of neuroevolution algorithms in reinforcement learning for self-driving cars,'' \emph{European Journal of Engineering Science and Technology}, vol.~2, no.~4, pp. 60--71, 2019.

\bibitem{kelly2021evolving}
S.~Kelly, T.~Voegerl, W.~Banzhaf, and C.~Gondro, ``Evolving hierarchical memory-prediction machines in multi-task reinforcement learning,'' \emph{Genetic Programming and Evolvable Machines}, vol.~22, pp. 573--605, 2021.

\bibitem{girgin2008feature}
S.~Girgin and P.~Preux, ``Feature discovery in reinforcement learning using genetic programming,'' in \emph{European conference on genetic programming}.\hskip 1em plus 0.5em minus 0.4em\relax Springer, 2008, pp. 218--229.

\bibitem{sehgal2019deep}
A.~Sehgal, H.~La, S.~Louis, and H.~Nguyen, ``Deep reinforcement learning using genetic algorithm for parameter optimization,'' in \emph{2019 Third IEEE International Conference on Robotic Computing (IRC)}.\hskip 1em plus 0.5em minus 0.4em\relax IEEE, 2019, pp. 596--601.

\bibitem{aydeniz2023novelty}
A.~A. Aydeniz, R.~Loftin, and K.~Tumer, ``Novelty seeking multiagent evolutionary reinforcement learning,'' in \emph{Proceedings of the Genetic and Evolutionary Computation Conference}, 2023, pp. 402--410.

\bibitem{zhu2021evolutionary}
S.~Zhu, F.~Belardinelli, and B.~G. Le{\'o}n, ``Evolutionary reinforcement learning for sparse rewards,'' in \emph{Proceedings of the Genetic and Evolutionary Computation Conference Companion}, 2021, pp. 1508--1512.

\bibitem{li2023race}
P.~Li, J.~Hao, H.~Tang, Y.~Zheng, and X.~Fu, ``Race: improve multi-agent reinforcement learning with representation asymmetry and collaborative evolution,'' in \emph{International Conference on Machine Learning}.\hskip 1em plus 0.5em minus 0.4em\relax PMLR, 2023, pp. 19\,490--19\,503.

\bibitem{chang2018genetic}
S.~Chang, J.~Yang, J.~Choi, and N.~Kwak, ``Genetic-gated networks for deep reinforcement learning,'' \emph{Advances in neural information processing systems}, vol.~31, 2018.

\bibitem{tang2021guiding}
Y.~Tang, ``Guiding evolutionary strategies with off-policy actor-critic.'' in \emph{AAMAS}, 2021, pp. 1317--1325.

\bibitem{simmons2019q}
R.~Simmons-Edler, B.~Eisner, E.~Mitchell, S.~Seung, and D.~Lee, ``Q-learning for continuous actions with cross-entropy guided policies,'' \emph{arXiv preprint arXiv:1903.10605}, 2019.

\bibitem{gao2023efficient}
M.~Gao, X.~Feng, H.~Yu, and X.~Li, ``An efficient evolutionary algorithm based on deep reinforcement learning for large-scale sparse multiobjective optimization,'' \emph{Applied Intelligence}, pp. 1--24, 2023.

\bibitem{tran2023two}
H.-L. Tran, L.~Doan, N.~H. Luong, and H.~T.~T. Binh, ``A two-stage multi-objective evolutionary reinforcement learning framework for continuous robot control,'' in \emph{Proceedings of the Genetic and Evolutionary Computation Conference}, 2023, pp. 577--585.

\bibitem{stork2021behavior}
J.~Stork, M.~Zaefferer, N.~Eisler, P.~Tichelmann, T.~Bartz-Beielstein, and A.~Eiben, ``Behavior-based neuroevolutionary training in reinforcement learning,'' in \emph{Proceedings of the Genetic and Evolutionary Computation Conference Companion}, 2021, pp. 1753--1761.

\bibitem{song2023rl}
Y.~Song, L.~Wei, Q.~Yang, J.~Wu, L.~Xing, and Y.~Chen, ``Rl-ga: A reinforcement learning-based genetic algorithm for electromagnetic detection satellite scheduling problem,'' \emph{Swarm and Evolutionary Computation}, vol.~77, p. 101236, 2023.

\bibitem{stork2019surrogate}
J.~Stork, M.~Zaefferer, T.~Bartz-Beielstein, and A.~Eiben, ``Surrogate models for enhancing the efficiency of neuroevolution in reinforcement learning,'' in \emph{Proceedings of the genetic and evolutionary computation conference}, 2019, pp. 934--942.

\bibitem{espositi2020gradient}
F.~Espositi and A.~Bonarini, ``Gradient bias to solve the generalization limit of genetic algorithms through hybridization with reinforcement learning,'' in \emph{Machine Learning, Optimization, and Data Science: 6th International Conference, LOD 2020, Siena, Italy, July 19--23, 2020, Revised Selected Papers, Part I 6}.\hskip 1em plus 0.5em minus 0.4em\relax Springer, 2020, pp. 273--284.

\bibitem{li2022reinforcement}
Y.~Li, ``Reinforcement learning in practice: Opportunities and challenges,'' \emph{arXiv preprint arXiv:2202.11296}, 2022.

\bibitem{rasouli2019autonomous}
A.~Rasouli and J.~K. Tsotsos, ``Autonomous vehicles that interact with pedestrians: A survey of theory and practice,'' \emph{IEEE transactions on intelligent transportation systems}, vol.~21, no.~3, pp. 900--918, 2019.

\bibitem{bai2023towards}
Y.~Bai, H.~Zhao, X.~Zhang, Z.~Chang, R.~J{\"a}ntti, and K.~Yang, ``Towards autonomous multi-uav wireless network: A survey of reinforcement learning-based approaches,'' \emph{IEEE Communications Surveys \& Tutorials}, 2023.

\bibitem{li2021meta}
J.-Y. Li, Z.-H. Zhan, K.~C. Tan, and J.~Zhang, ``A meta-knowledge transfer-based differential evolution for multitask optimization,'' \emph{IEEE Transactions on Evolutionary Computation}, vol.~26, no.~4, pp. 719--734, 2021.

\bibitem{aleti2016systematic}
A.~Aleti and I.~Moser, ``A systematic literature review of adaptive parameter control methods for evolutionary algorithms,'' \emph{ACM Computing Surveys (CSUR)}, vol.~49, no.~3, pp. 1--35, 2016.

\bibitem{zhan2019adaptive}
Z.-H. Zhan, Z.-J. Wang, H.~Jin, and J.~Zhang, ``Adaptive distributed differential evolution,'' \emph{IEEE transactions on cybernetics}, vol.~50, no.~11, pp. 4633--4647, 2019.

\bibitem{hochreiter1997long}
S.~Hochreiter and J.~Schmidhuber, ``Long short-term memory,'' \emph{Neural computation}, vol.~9, no.~8, pp. 1735--1780, 1997.

\bibitem{shi2022reinforcement}
H.~Shi, B.~Zhou, H.~Zeng, F.~Wang, Y.~Dong, J.~Li, K.~Wang, H.~Tian, and M.~Q.-H. Meng, ``Reinforcement learning with evolutionary trajectory generator: A general approach for quadrupedal locomotion,'' \emph{IEEE Robotics and Automation Letters}, vol.~7, no.~2, pp. 3085--3092, 2022.

\bibitem{weiss2016survey}
K.~Weiss, T.~M. Khoshgoftaar, and D.~Wang, ``A survey of transfer learning,'' \emph{Journal of Big data}, vol.~3, no.~1, pp. 1--40, 2016.

\bibitem{ajao2023secure}
L.~A. Ajao and S.~T. Apeh, ``Secure edge computing vulnerabilities in smart cities sustainability using petri net and genetic algorithm-based reinforcement learning,'' \emph{Intelligent Systems with Applications}, vol.~18, p. 200216, 2023.

\bibitem{hong2021federated}
J.~Hong, Z.~Zhu, S.~Yu, Z.~Wang, H.~H. Dodge, and J.~Zhou, ``Federated adversarial debiasing for fair and transferable representations,'' in \emph{Proceedings of the 27th ACM SIGKDD Conference on Knowledge Discovery \& Data Mining}, 2021, pp. 617--627.

\bibitem{petrovic2022fair}
A.~Petrovi{\'c}, M.~Nikoli{\'c}, S.~Radovanovi{\'c}, B.~Deliba{\v{s}}i{\'c}, and M.~Jovanovi{\'c}, ``Fair: Fair adversarial instance re-weighting,'' \emph{Neurocomputing}, vol. 476, pp. 14--37, 2022.

\bibitem{finn2017model}
C.~Finn, P.~Abbeel, and S.~Levine, ``Model-agnostic meta-learning for fast adaptation of deep networks,'' in \emph{International conference on machine learning}.\hskip 1em plus 0.5em minus 0.4em\relax PMLR, 2017, pp. 1126--1135.

\bibitem{elsayed2012improved}
S.~M. Elsayed, R.~A. Sarker, and D.~L. Essam, ``An improved self-adaptive differential evolution algorithm for optimization problems,'' \emph{IEEE Transactions on Industrial Informatics}, vol.~9, no.~1, pp. 89--99, 2012.

\bibitem{peng2023reinforcement}
L.~Peng, Z.~Yuan, G.~Dai, M.~Wang, and Z.~Tang, ``Reinforcement learning-based hybrid differential evolution for global optimization of interplanetary trajectory design,'' \emph{Swarm and Evolutionary Computation}, p. 101351, 2023.

\bibitem{niu2020decade}
S.~Niu, Y.~Liu, J.~Wang, and H.~Song, ``A decade survey of transfer learning (2010--2020),'' \emph{IEEE Transactions on Artificial Intelligence}, vol.~1, no.~2, pp. 151--166, 2020.

\bibitem{lian2019towards}
D.~Lian, Y.~Zheng, Y.~Xu, Y.~Lu, L.~Lin, P.~Zhao, J.~Huang, and S.~Gao, ``Towards fast adaptation of neural architectures with meta learning,'' in \emph{International Conference on Learning Representations}, 2019.

\bibitem{zhang2023coordinated}
T.~Zhang, L.~Yu, D.~Yue, C.~Dou, X.~Xie, and L.~Chen, ``Coordinated voltage regulation of high renewable-penetrated distribution networks: an evolutionary curriculum-based deep reinforcement learning approach,'' \emph{International Journal of Electrical Power \& Energy Systems}, vol. 149, p. 108995, 2023.

\bibitem{yang2023evolutionary}
X.~Yang, N.~Chen, S.~Zhang, X.~Zhou, L.~Zhang, and T.~Qiu, ``An evolutionary reinforcement learning scheme for iot robustness,'' in \emph{2023 26th International Conference on Computer Supported Cooperative Work in Design (CSCWD)}.\hskip 1em plus 0.5em minus 0.4em\relax IEEE, 2023, pp. 756--761.

\bibitem{malandri2023model}
L.~Malandri, F.~Mercorio, M.~Mezzanzanica, and A.~Seveso, ``Model-contrastive explanations through symbolic reasoning,'' \emph{Decision Support Systems}, p. 114040, 2023.

\bibitem{porebski2022evaluation}
S.~Porebski, ``Evaluation of fuzzy membership functions for linguistic rule-based classifier focused on explainability, interpretability and reliability,'' \emph{Expert Systems with Applications}, vol. 199, p. 117116, 2022.

\bibitem{shao2023effect}
X.~Shao, H.~Wang, X.~Zhu, F.~Xiong, T.~Mu, and Y.~Zhang, ``Effect: Explainable framework for meta-learning in automatic classification algorithm selection,'' \emph{Information Sciences}, vol. 622, pp. 211--234, 2023.

\bibitem{min2021recent}
B.~Min, H.~Ross, E.~Sulem, A.~P.~B. Veyseh, T.~H. Nguyen, O.~Sainz, E.~Agirre, I.~Heintz, and D.~Roth, ``Recent advances in natural language processing via large pre-trained language models: A survey,'' \emph{ACM Computing Surveys}, 2021.

\bibitem{chang2023examining}
E.~Y. Chang, ``Examining gpt-4: Capabilities, implications and future directions,'' in \emph{The 10th International Conference on Computational Science and Computational Intelligence}, 2023.

\bibitem{zhong2023adapter}
S.~Zhong, Z.~Huang, W.~Wen, J.~Qin, and L.~Lin, ``Sur-adapter: Enhancing text-to-image pre-trained diffusion models with large language models,'' in \emph{Proceedings of the 31st ACM International Conference on Multimedia}, 2023, pp. 567--578.

\end{thebibliography}

\vfill

\end{document}